\documentclass[sigconf]{acmart}
\AtBeginDocument{%
  }

\copyrightyear{2026}
\acmYear{2026}
\setcopyright{cc}
\setcctype{by}
\acmConference[MM '26]{Proceedings of the 34th ACM International Conference on Multimedia}{November 10--14, 2026}{Rio de Janeiro, Brazil}
\acmBooktitle{Proceedings of the 34th ACM International Conference on Multimedia (MM '26), November 10--14, 2026, Rio de Janeiro, Brazil}
\acmDOI{10.1145/3767308.3836037}
\acmISBN{979-8-4007-2213-4/2026/11}

\usepackage{algorithm}
\usepackage{algorithmic}
\usepackage{epigraph}
\usepackage{svg}
\usepackage{multirow}
\usepackage{subcaption}
\usepackage{placeins}
\usepackage{longtable}
\begin{document}

\title{Zero-MELO: Test-Time Evidence Calibration with Multimodal LLMs for Zero-Shot Micro-Gesture Recognition}

\author{Chengyan Wang}
\orcid{0009-0009-2595-1316}
\affiliation{%
  \institution{CMVS, University of Oulu}
  \city{Oulu}
  \country{Finland}
}
\email{chengyan.wang@oulu.fi}

\author{Hanliang Xie}
\orcid{0009-0002-1093-2197}
\affiliation{%
  \institution{Peking University}
  \city{Beijing}
  \country{China}}
\email{xiehl@stu.pku.edu.cn}

\author{Yueyi Yang}
\orcid{0000-0002-4948-1058}
\affiliation{%
  \institution{CMVS, University of Oulu}
  \city{Oulu}
  \country{Finland}
}
\email{yueyi.yang@oulu.fi}


\author{Haoyu Chen}
\authornote{Corresponding author.}
\orcid{0000-0003-3267-2664}
\affiliation{%
  \institution{CMVS, University of Oulu}
  \city{Oulu}
  \country{Finland}
}
\email{chen.haoyu@oulu.fi}

\renewcommand{\shortauthors}{Chengyan Wang, Hanliang Xie, Yueyi Yang, and Haoyu Chen}


\begin{abstract}   
  While Multimodal Large Language Models (MLLMs) excel in general video understanding, their capability in fine-grained and motion-centric tasks remains limited. This limitation is particularly critical in micro-gesture recognition (MGR), where \textbf{micro-gestures} (MG)—subtle, short-duration, and spatially localized human movements—serve as key discriminative signals for implicit affective analysis, yet are easily neglected following common prompting practices. Although MGR has been intensively studied by many discriminative approaches, the use of MLLMs on MGR is underexplored, with notably poor performance. We hypothesize that the motion-sensitive representation ability of MLLMs is constrained by its inherent single-pass forward inference, which can be substantially enhanced through carefully designed test-time guidance. Motivated by this, building on our prior findings regarding temporal insensitivity in Video LLMs, we diagnose zero-shot MGR errors in the Negative Log-Likelihood (NLL) space. We observe that MLLMs suffer from two bottlenecks: \textbf{1) insufficient localized evidence} and \textbf{2) severe score biases} driven by language and motion-agnostic appearances. Thus, we propose a novel test-time evidence calibration framework that improves both reasoning details and prediction reliability. Specifically, we introduce a tree search mechanism to progressively acquire localized, fine-grained visual evidence, coupled with a test-time calibration module to mitigate score biases. The multi-cue fusion module then integrates evidence from multiple cues without relying on a single cue for final prediction. Our framework achieves a mean-class accuracy of $26.84\%$ on iMiGUE and $22.10\%$ on MA-52, significantly outperforming the Qwen2.5-VL baseline that produces $16.15\%$ and $10.20\%$, respectively. The code will be available at https://zero-melo.github.io/Zero-MELO.
\end{abstract} 






\maketitle
\begin{figure}[h]
    \centering
    \includegraphics[width=\linewidth]{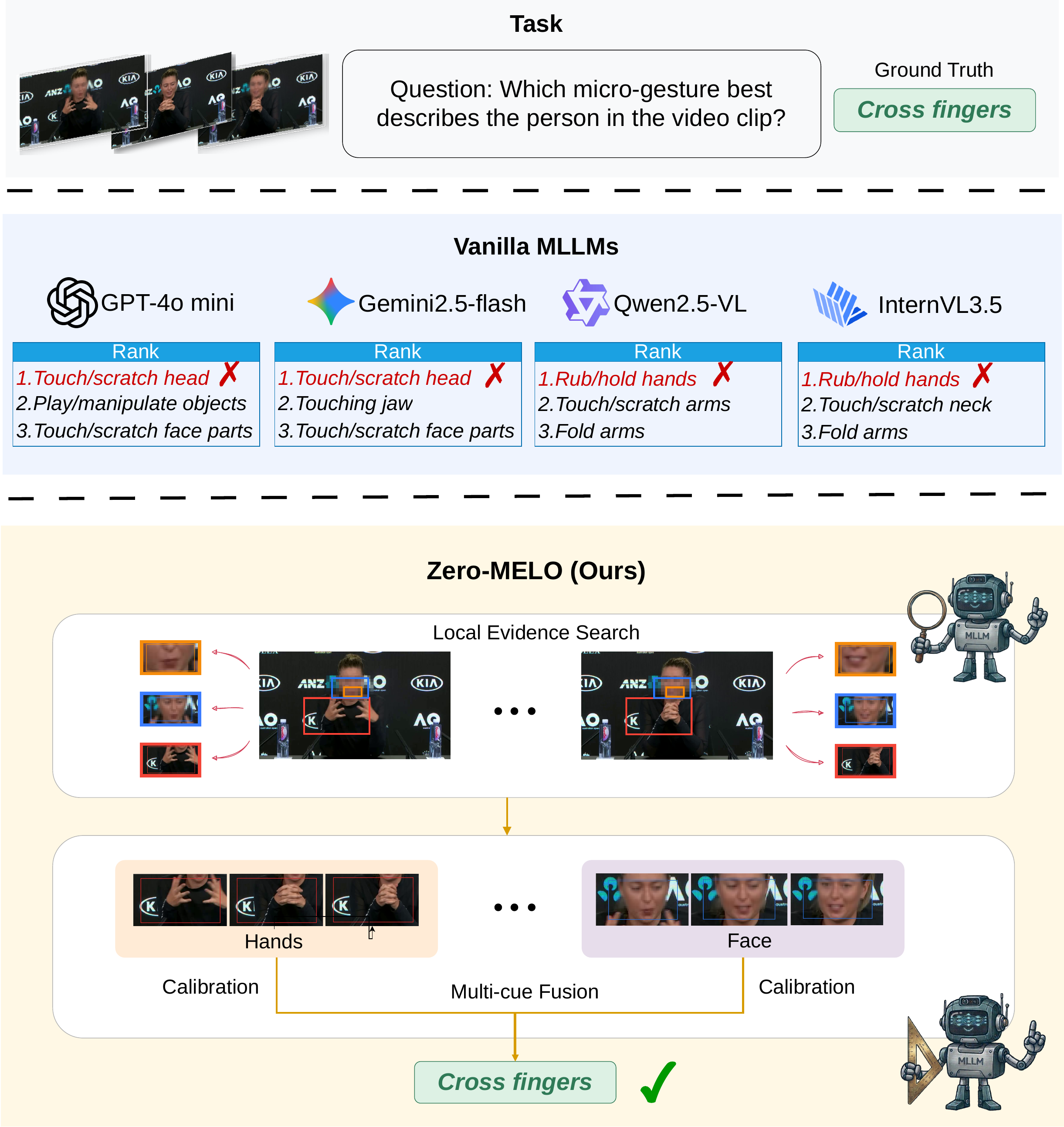}
    \caption{Comparison between existing methods and Zero-MELO. Given the input video clips and question, the target is to predict the correct label in MGR. Compared to the single-pass forward inference of other MLLMs, Zero-MELO can utilize the relevant local evidence and predict the correct label via calibration at inference time. }
    \label{fig:1}
\end{figure}  
\section{Introduction}
     
Micro-gestures (MG) are subtle, short-duration, and spatially localized human movements that serve as key indicators of underlying behavioral and affective states \cite{BodyCuesNotFacialExpressions}, which has drawn increasing research attention in recent years. Compared with general human activities \cite{nturgbd, pkummd, kinetics400, nturgbd120}, MG shows finer granularity in both temporal and spatial domains, making its reliable recognition difficult \cite{motionmatters}. The field has progressed rapidly with benchmarks of increasing scale and coverage, including iMiGUE \cite{imigue, kakouros2026imigue}, SMG \cite{smg}, BBSI \cite{BBSI}, MA-52 \cite{ma52,li2026mac}, MMA-52 \cite{mma52}, and iMiGUE-3K \cite{imigue3k}, and supervised models have achieved strong in-domain performances \cite{MAC2024_1st,mmgesture,motionmatters,Prototpyical_Calibrating}. 

Nevertheless, micro-gesture recognition (MGR) remains fundamentally challenging due to several intrinsic factors. First, MGs exhibit substantial cross-subject variability, where subtle motion patterns are heavily entangled with individual-specific appearance and motion styles. Second, MG categories typically follow a long-tailed distribution, where rare and fine-grained gestures lack sufficient training samples, leading to biased decision boundaries in supervised settings. Third, real-world MG scenarios continuously introduce \textbf{unseen} or \textbf{novel} gesture semantics, which cannot be exhaustively covered by predefined label spaces. As a result, conventional fully supervised MGR paradigms, which rely on fixed label sets and sufficient annotated data, struggle to scale to realistic settings. These limitations naturally motivate \textbf{zero-shot MGR}, where models are required to recognize fine-grained and potentially novel micro-gestures without additional task-specific training.

Meanwhile, large language models (LLMs) and MLLMs have demonstrated great zero-shot and few-shot capability across various tasks \cite{GPT-3, qwen3, internvl35, glm45v, lian2025affectgpt}, including both general-purpose \cite{mmbench} and specific downstream tasks \cite{MME, MME-RealWorld, v_guided, yuan2026motor, bai2026exploring, lian2024ov, lian2025affectgptr1,lian2025emoprefer}, which naturally motivates the zero-shot MGR. Despite the broad transfer capability shown in MLLMs, their capability on fine-grained motion-centric video understanding remains limited \cite{motionsight,yang2026saynext}. Our preliminary study shows that \textit{directly applying MLLMs} on MGR leads to poor performance in both supervised and zero-shot settings (see more in Sect.~\ref{Sec:Anlaysis_of_failure}).


Thus, in this work, we revisit this discrepancy and ask a fundamental question: \textbf{Why do MLLMs struggle to capture subtle, motion-critical cues in micro-gesture recognition under standard prompting settings?} To better understand this problem, we start from standard prompt-based free-form generation, and further analyze MLLM predictions from both free-form generation and Negative Log-Likelihood (NLL)-based label scoring that ranks candidate labels by NLL and selects the minimum-NLL label. Our prior qualitative studies have argued that MLLMs can recognize obvious micro-gestures on iMiGUE but often fail on subtle ones (e.g., when the movement is extremely subtle), suggesting that common general prompting is insufficient for preserving decisive local motion cues in zero-shot MGR \cite{gpt_as_psychologist}. Furthermore, the NLL-based analysis quantitatively reveals a second issue underlying why MLLMs fail in MGR, namely a strong prior-driven bias in general MLLMs. Specifically, we find that \textbf{when visual input is removed, the model continues to produce confident predictions, revealing a language prior bias}, indicating its tendency to rely on semantic expectations. This is consistent with existing research findings \cite{contrastive_decoding_in_test1,contrastive_decoding_in_test3}. Conversely, when using static (rank-1) visual cues that preserve appearance but remove motion, the model exhibits an appearance bias, which is globally weaker but can be locally stronger for certain classes. These two biases heavily affect the behavior of MLLMs, especially under long-tailed label distributions. Based on this analysis, we show that reliable zero-shot MGR requires both better \textbf{extraction of localized motion evidence} and \textbf{bias-aware score/evidence calibration}.



Motivated by these insights, we propose to address the above bottlenecks by expanding \textit{test-time} reasoning to unlock the latent capability of MLLMs for zero-shot MGR. Specifically, instead of treating the model’s first-pass response as final, we introduce a more elaborate inference procedure that searches for localized motion evidence, aggregates clues across spatial-temporal regions rather than relying on a single cue, and recalibrates predictions against dominant semantic priors. Thus, we propose \textbf{Zero-MELO}, a training-free MLLM framework for zero-shot MGR via test-time refinement. It incorporates a cue-guided tree search mechanism to acquire the zoomed local evidence from informative body cues, a dual test-time calibration module to mitigate the prior-driven score bias, and a cue-conditioned evidence fusion module to aggregate NLL scores across different cues to predict the final label. The whole framework comparison with conventional MLLMs is shown in Fig.~\ref{fig:1}. Under the Qwen2.5-VL \cite{qwen25vl} backbone, our framework improves mean-class accuracy from $16.15\%$ to $26.84\%$ on iMiGUE and from $10.20\%$ to $22.10\%$ on MA-52.

\noindent \textbf{In summary, our contributions can be concluded as follows:}

(1) We revisit the failure of MLLMs in micro-gesture recognition and show that the bottleneck is not solely due to insufficient representation ability of MLLMs, but is largely caused by insufficient localized evidence and severe score biases driven by language- and motion-agnostic appearances.

(2) We introduce the idea of expanding test-time reasoning in MLLMs for zero-shot MGR, demonstrating that enhancing evidence exploration and aggregation at inference time can effectively reveal the latent capability of MLLMs.

(3) We propose Zero-MELO, a training-free MLLM framework that couples cue-shared zoom search with dual-reference calibration to address missing local evidence and prior-driven score bias in zero-shot MGR with a cue-conditioned evidence fusion strategy, enabling robust multi-cue recognition under shared context.

(4) Extensive experiments show that Zero-MELO consistently improves over standard MLLM inference, suggesting that the limitations of MLLMs on micro-gesture recognition are highly related to insufficient localized evidence and severe biases.


\section{Related Work}
\textbf{Zero-shot Micro-Gesture Recognition.} Existing MGR methods are predominantly supervised and learn dataset-specific spatiotemporal representations from RGB clips \cite{3d_cnn_mgr, posec3d}, skeleton \cite{stgcn, hand_crafted_stgcn, blockgcn} sequences, cross-modality fusion \cite{GDN_MG,clip-mg,mmgesture} or the joint modeling of spatial structure and temporal dynamics \cite{MAC2024_1st, motionmatters, online_mgr_data_aug_spatial_temporal_attention,kmanet, HyperGcnTransformer}. Despite this progress, these methods typically depend on labeled training data and are not directly applicable in a zero-shot setting.

Existing MLLM studies on MGR-related datasets mainly focus on affective understanding or semantic analysis, rather than action-level MGR \cite{deemo, identity_free_emotion_llm, shang2025cross}. Moreover, previous qualitative evidence suggests that global-view prompting can recognize obvious MG yet often fails on subtle ones in iMiGUE \cite{gpt_as_psychologist}. These observations indicate that directly applying a generic MLLM to MGR is insufficient, but they do not provide a quantitative diagnosis of why zero-shot MGR fails. Different from these works, we study the predefined-set zero-shot MGR based on MLLM label-scoring formulation, and quantitatively analyze its two core bottlenecks: insufficient localized evidence and reference-conditioned bias in label scores.

\noindent\textbf{Test-Time Inference in MLLMs.}
Test-time inference in LLMs has become an important research direction \cite{test_time_inference_2, test_time_inference_3, test_time_inference_4}. A major line of recent work addresses hallucination and prior-driven errors in LLMs/MLLMs at test time. Prior work mainly mitigates hallucination and prior-driven errors through contrastive decoding \cite{contrastive_decoding_in_test1, contrastive_decoding_in_test2, contrastive_decoding_in_test3, VCD}, attention-space or hidden-state interventions \cite{woo-etal-2025-dont, ibd, devils_in_middle_layers, agla, clearsight, modular_attribution_and_intervention, vti_decoding, bidirectional_hidden_state_intervention}, dynamic decoding \cite{octpus}, leveraging external generation models \cite{CoFi-Dec}.
Compared to these image-based methods mentioned above, video-specific variants have been explored to suppress temporally insensitive responses and improve temporal reasoning \cite{SID, TCD, improve_temporal_reasoning}.


Although highly relevant, these methods primarily intervene in token-level decoding or hidden-state dynamics in open-ended generation settings. In contrast, we consider the MGR with a predefined label set. Accordingly, rather than intervening in generation-time decoding, we reformulate the idea as bias calibration on label NLLs, where the calibration procedure directly adjusts label scores rather than generation-time token logits. This ensures all calibrated predictions remain valid and allows the correction signal to be naturally integrated with multi-cue evidence fusion.

\noindent\textbf{MLLMs with Fine-grained Visual Search.}
Visual encoders trained through contrastive learning, such as CLIP \cite{clip} and SigLip \cite{siglip}, often discard fine-grained semantics \cite{clip, siglip, LLava-onevision-1.5}.
To address this issue, prior work augments perception with local crops \cite{dc2, v_guided}, search-based zooming \cite{zoomeye, zoom-refine}, model-internal focusing mechanisms \cite{vicrop, Focus, adafocus}, retrieval-assisted perception \cite{RAP}, or training-based active perception \cite{deepeyes, Thyme, deepeyesv2, zooming_without_zooming, Video-thinker}. These methods broadly follow a ``thinking with images'' paradigm \cite{thinking-with-image}.

Our setting is fundamentally different due to the video-based and predefined label list nature of MGR. Hence, the goal is not to find arbitrary question-relevant regions, but to acquire discriminative local evidence for classification. Moreover, unlike instance-specific search targets, our search is cue-shared across labels, allowing localized evidence to be reused across multiple actions. Besides, existing methods typically use one selected crop as auxiliary evidence. In contrast, zero-shot MGR does not provide an oracle cue prior: the challenge is not only localized search, but also cross-cue decision under branch-specific score distortion and shared global context.

\begin{figure*}[h]
    \centering
    \includegraphics[width=\linewidth]{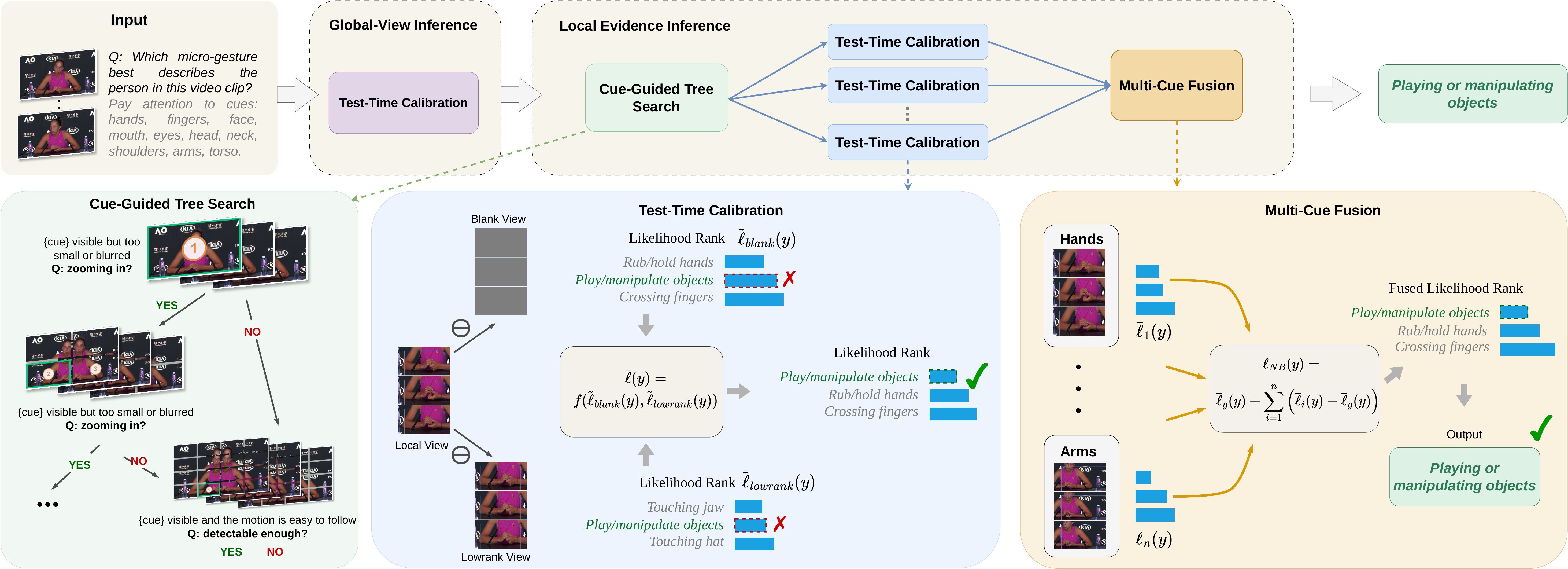}
    \caption{The overall algorithm of Zero-MELO. It mainly consists of three key modules: the cue-guided search (Sec.~\ref{Sec:tree_search}), the test-time calibration (Sec.~\ref{Sec:method_calibration}), and the multi-cue fusion (Sec.~\ref{Sec:method_cue_fusion}). The global video input will first be fed to the MLLM for the global inference. Then, the model will acquire local evidence via the cue-guided tree search. Next, the test-time calibration will be conducted, and the calibrated scores will be yielded. Based on the global and local calibrated scores, the multi-cue fusion contributes to the final prediction.}
    \label{fig:2}
\end{figure*}  
\section{Methodology}
\label{Sec:Methodology}
\subsection{Problem Formulation}
\label{Sec:ProblemForm}
The zero-shot MGR with an MLLM is defined as follows: given an original video clip $g$ and a prompt $t$, the model predicts a corresponding MG label $y \in \mathcal{Y}$,
\begin{equation}
y^* = \arg\max_{y \in \mathcal{Y}} p_{\mathcal{M}}(y \mid g,t),
\end{equation}
where $p_{\mathcal{M}}(y \mid g,t)$ denotes the label-level prediction for class $y$ under a specific inference scheme, and $p_{\mathcal{M}}$ represents the MLLM with parameters $\mathcal{M}$.

To avoid notation ambiguity, we distinguish three types of visual input. We use $g$ for the original/global video clip, $z$ for a localized cue view derived from $g$ by our method, and $x$ for a generic visual input when a formulation is shared across input types. Accordingly, $x$ can instantiate the global clip $g$, a localized view $z$, or a joint input such as $(g, z)$.

\noindent\textbf{Prompt-based free-form generation vs. NLL-based label scoring.} An intuitive approach to zero-shot MGR is the common general prompting setting referred to as prompt-based free-form generation in this work, where the MLLM directly produces predictions conditioned on a candidate label list. While this paradigm serves as a competitive baseline following common prompting practices, it \textit{only exposes the final generated output}, without access to internal token-level likelihoods or logits. As a result, it cannot provide a calibrated and comparable score distribution over the full label space. This limitation is particularly restrictive for our framework, which requires label-wise score vectors to diagnose bias and perform subsequent correction. Moreover, free-form generation is inherently misaligned with a predefined label inventory, especially in closed-source LLMs where internal scoring signals are not accessible. See Supplementary Material A (SM A) for misaligned failure cases.

To address these issues, we choose a different technical path that reformulates zero-shot MGR as a classification problem. Specifically, we compute a score for each candidate label by evaluating the teacher-forced negative log-likelihood (NLL) of its verbalized form. This enables consistent and comparable scoring across all labels, while leveraging the model’s internal probabilistic signals for more transparent decision-making. Details are presented in Sec.~\ref{Sec:method_calibration}.

\subsection{Overview of Zero-MELO}
The overall framework of Zero-MELO is illustrated in Fig.~\ref{fig:2}. The data flow of Zero-MELO starts from the global input video MG clip and prompt $(g, t)$, where MLLM computes label-wise scores $s_g(y)$ or equivalently NLL $\ell_g(y)$ over the set of labels $\mathcal{Y}$. To address missing fine-grained evidence, we propose a \textbf{\textit{1) cue-guided tree search} }that generates local views $\{z_i\}_{i=1}^n$ from semantic cues $c_i$, forming paired inputs $x_i = (g, z_i)$, each producing scores $s_{x_i}(y)$. To mitigate prior bias, \textbf{\textit{2) test-time calibration}} introduces the calibration inputs of $x_b$ (blank) and $x_r$ (lowrank), yielding calibrated NLL $\tilde{\ell}_{x_b}(y)$ and $\tilde{\ell}_{x_r}(y)$, which are further combined into aggregated calibration energies $\bar{\ell}_x(y)$. Finally, \textbf{\textit{3) multi-cue fusion}} aggregates global and local evidence into a unified score $\ell_{\text{fuse}}(y)$, and the final prediction is obtained as $y_{\text{final}}$. Below, we introduce the Zero-MELO following the data flow in detail.


\subsection{Cue-Guided Tree Search}
\label{Sec:tree_search}

As revealed by our analytical experiments in Sec.~\ref{Sec:Anlaysis_of_failure}, we can see that a key failure mode of MLLMs in MGR lies in their improper cue allocation: under common prompting practices, the model often attends to irrelevant regions while neglecting informative, motion-critical cues, leading to insufficient discriminative evidence. To address this, inspired by the general zoom-search idea of some existing works~\cite{zoomeye, vicrop, zoom-refine}, we design a novel technique called \emph{Cue-Guided Tree Search} customized for video-based MGR. 



Given an input video, we first select a motion-salient temporal window and uniformly sample $T$ frames from it to form the global clip $g$. Rather than searching on a single frame, for each cue $c_i$ (e.g., "pay attention to the lips"), we perform cue-guided search on several anchor frames $\mathcal{A}=\{a_1,\dots,a_m\}$ sampled from the clip:
\begin{equation}
\mathcal{B}_{i,a}=\mathrm{search}(g_a,c_i), \qquad a\in\mathcal{A},
\end{equation}
where $\mathcal{B}_{i,a}$ denotes the high-confidence proposals returned on anchor $a$. This anchor-based design improves temporal stability for transient motions and avoids relying on a single snapshot.

The retained anchor-wise proposals are then aggregated into one temporally stable region,
\begin{equation}
b_i=\mathrm{Agg}\Big(\bigcup_{a\in\mathcal{A}}\hat{\mathcal{B}}_{i,a}\Big),
\end{equation}
where $\hat{\mathcal{B}}_{i,a}\subseteq\mathcal{B}_{i,a}$ is a small retained set, and $\mathrm{Agg}(\cdot)$ performs robust cross-anchor aggregation by suppressing inconsistent proposals and unioning the remaining ones. If the aggregated region is unreliable or becomes excessively large, we conservatively fall back to the root region. The final cue-conditioned local view is obtained by applying the same crop to all sampled frames:
\begin{equation}
z_i=\mathrm{CropResize}(g,b_i).
\end{equation}

Compared with generic zooming on a single image, this design adapts cue search to video MGR by combining motion-aware temporal selection, anchor-based localization, and robust cross-anchor aggregation. In practice, the resulting cue-conditioned views are reused for all-label NLL scoring and later fused by the multi-cue inference module. Furthermore, several efficiency-oriented implementation optimizations are employed, including caching and batched inference, without changing the scoring objective. The implementation details are discussed in the SM B.

\subsection{NLL-based Test-Time Calibration}
\label{Sec:method_calibration}
Our analysis in Sec.~\ref{Sec:Anlaysis_of_failure} shows that the failure of vanilla MLLMs in zero-shot MGR is not only caused by missing local evidence, but also by severe bias in label prediction. For instance, the model tends to collapse to a few dominant labels. These observations suggest that raw label ranking is substantially affected by prior-driven bias, rather than being determined by the visual evidence. As a result, directly using the raw label scores can lead to unreliable predictions.

To address this issue, we introduce an NLL-based test-time calibration module. The core idea is to compare the NLL of an input view $x$ with two controlled calibration variants: a blank calibration input $x_b$ obtained by masking the video input and a lowrank calibration input $x_r$ (motion-based calibration) obtained by repeating a single frame across time. The blank calibration input suppresses visual content and mainly reflects language prior, while the lowrank calibration input preserves static appearance but removes motion information.

Let $\mathbf{v}(y)$ denote the verbalized label string. For a general input view $x$, we define the per-label NLL as
\begin{equation}
\ell_x(y) = -\frac{1}{|\mathbf{v}(y)|} \log p_{\mathcal{M}}(\mathbf{v}(y)\mid x,t),
\end{equation}
with token-level factorization under teacher forcing. Hence, the prediction is
\begin{equation}
y^* = \arg\min_{y \in \mathcal{Y}} \ell_x(y).
\end{equation}

We further define the induced label distribution:
\begin{equation}
\hat p_x(y) = \frac{\exp(-\ell_x(y))}{\sum_{y'} \exp(-\ell_x(y'))}.
\end{equation}

We then perform calibration in NLL space by
\begin{equation}
\tilde{\ell}_{x_b}(y)\triangleq \ell_{x}(y)-\ell_{x_{b}}(y),
\qquad
\tilde{\ell}_{x_r}(y)\triangleq \ell_{x}(y)-\ell_{x_{r}}(y).
\end{equation}
The calibrated scores are aggregated by:
\begin{equation}
\bar{\ell}_{x}(y)
\triangleq
-\tau_d \log \!\left(
\pi_x \exp\!\left(-\frac{\tilde{\ell}_{x_b}(y)}{\tau_d}\right)
+
(1-\pi_x)\exp\!\left(-\frac{\tilde{\ell}_{x_r}(y)}{\tau_d}\right)
\right),
\label{eq:dual_calibrated_energy}
\end{equation}
where $\tau_d>0$ is a temperature parameter, and $\pi_x\in(0,1)$ controls the relative contribution of the blank and lowrank references. The resulting calibrated energy corrects prior-driven score distortions and provides comparable global and cue-conditioned
branch energies for subsequent fusion, making calibration a
diagnosis-driven core component rather than auxiliary
post-processing. See the full derivation in SM C.

\subsection{Multi-Cue Fusion}
\label{Sec:method_cue_fusion}
Given the cue-guided zoomed-in views $z_{1:n}$ extracted from the global clip $g$ in the tree search stage, we present the multi-cue fusion module that fuses multiple local evidences into a single label distribution over the candidate set $\mathcal{Y}$.

For each cue view $z_i$, a paired input $x_i \triangleq (g, z_i)$ is constructed and scored by the MLLM using the same teacher-forced objective in Sec.~\ref{Sec:method_calibration}, yielding per-label loss $\ell_{x_i}(y)$, and the induced distribution $\hat p_{x_i}(y)$. The module is designed to derive an aggregation rule fusing $\{\hat p_{x_i}(y)\}_{i=1}^{n}$ to predict the $y_{final}$ instead of selecting one cue branch.

The product of experts \cite{poe} is adopted to model the joint conditional probability,
\begin{equation}
    p(y\mid x_{1},\dots x_{n})\approx \prod_{i=1}^n p(y\mid x_{i}).
    \label{eq:poe}
\end{equation}

However, simply adopting products of $p(y \mid x_{i})$ will lead to the problem that the information of the global view will accumulate repeatedly because $\{x_{i}\mid i=1,\dots, n\}$ share the same global view. To address this issue, 
we adopt the Naive Bayes approximation \cite{naive_bayesian}, (see detailed derivation in SM D)
with terms independent of $y$ (e.g., $\prod_i p(z_i\mid g)$) and propose an ideal posterior factorization:

\begin{equation}
    p(y\mid g, z_{1:n}) \propto p(y\mid g)\prod_{i=1}^{n}\frac{p(y\mid g, z_{i})}{p(y\mid g)}=p(y\mid g)^{1-n}\prod_{i=1}^{n}p(y\mid g,z_{i}).
    \label{eq:final}
\end{equation}

\paragraph{\textbf{Implementation in Calibrated NLL Space}}
Eq.~\eqref{eq:final} gives an ideal posterior factorization under the conditional independence approximation. In practice, however, cue-conditioned branches are neither independent nor equally reliable; their rankings are not directly comparable, and the set of cues is label-dependent. Therefore,  Eq.~\eqref{eq:final} is retained as the probabilistic motivation, while instantiating the final inference rule in a calibrated energy space.

Using the notation in Sec.~\ref{Sec:method_calibration}, for the global branch $g$, the calibrated losses defined previously are used
$\tilde{\ell}_{g_b}(y)$ and $\tilde{\ell}_{g_r}(y)$. For the $i$-th paired branch
$x_i=(g,z_i)$, we abbreviate $\bar{\ell}_{(x_i)_b}(y)$ by $\bar{\ell}_{i,b}(y)$.

Hence, $\bar{\ell}_{x}(y)$ is a temperature-smoothed lower envelope of the two bias calibration energies, and serves as the branch-wise energy used in the subsequent fusion.

Let
\begin{equation}
\bar{p}_{x}(y)
\triangleq
\frac{\exp(-\bar{\ell}_{x}(y))}
{\sum_{y'\in\mathcal{Y}}\exp(-\bar{\ell}_{x}(y'))}
\label{eq:branch_surrogate_posterior}
\end{equation}
be the surrogate posterior induced by $\bar{\ell}_{x}(y)$.
Substituting Eq.~\eqref{eq:branch_surrogate_posterior} into the factorization in
Eq.~\eqref{eq:final} yields, up to a Naive-Bayesian (NB) label-independent constant,
\begin{equation}
\ell_{\mathrm{NB}}(y)
\doteq
\sum_{i=1}^{n}\bar{\ell}_{i,b}(y) - (n-1)\bar{\ell}_{g}(y)
=
\bar{\ell}_{g}(y)
+
\sum_{i=1}^{n}\bigl(\bar{\ell}_{i,b}(y)-\bar{\ell}_{g}(y)\bigr),
\label{eq:raw_residual_energy}
\end{equation}
which shows that each cue branch contributes a residual correction relative to the shared global branch.

However, Eq.~\eqref{eq:raw_residual_energy} is not yet suitable as the final rule, because not all cues are semantically relevant to every label. We propose a refined multi-cue fusion strategy, called \textbf{Label-Conditioned Normalized Fusion} (please see the detailed derivation in SM D).

Let $\mathcal{I}(y)=\{i : c_i\in\mathcal{C}(y)\}$ denote the compatible cue set, where $\mathcal{C}(y)$ is defined once from label semantics and body-part association (See full mapping in SM E.1), and $w_i=\lambda_{\phi(c_i)}$ the family weight. We fuse global and local evidence as
\begin{equation}
\ell_{\mathrm{fuse}}(y)
=
\frac{
\bar{\ell}_{g}(y)
+
\sum_{i\in\mathcal{I}(y)} w_i\,\bar{\ell}_{i}(y)
}{
1+\sum_{i\in\mathcal{I}(y)} w_i
}.
\end{equation}
The final prediction is
\begin{equation}
y_{\mathrm{final}}=\arg\min_{y\in\mathcal{Y}} \ell_{\mathrm{fuse}}(y).
\end{equation}

\section{Experiments}
\subsection{Experimental Setup}
\label{Sec:exp_setup}
\noindent\textbf{Datasets.} In the experiments, we evaluate our method on the iMiGUE \cite{imigue} test set and MA-52 \cite{ma52} validation set as they align with our clip-level MGR setting, while other available datasets follow different or derivative protocols. The primary MLLM backbone is Qwen2.5-VL-7B \cite{qwen25vl}, unless
otherwise specified.
For iMiGUE, experiments are conducted on a single NVIDIA A100 GPU with 40GB memory, whereas MA-52 experiments are conducted on AMD MI250X GPUs due to the larger size. See more details of experimental configurations, including the values of parameters and compatible cues used in multi-cue fusion, in SM E.1.

\noindent\textbf{Cue--label associations.}
The label-conditioned cue associations are generated once by an LLM based on label semantics and anatomical relevance, and are then fixed for all samples. No sample-specific annotation or ground-truth label is used to construct or select the cue associations during inference. The generation prompt and complete associations are provided in the supplementary material.

\noindent\textbf{Evaluation.} Since MG datasets are strongly long-tailed, to ensure fair evaluation, we report mean-class Top-1 (mCA@1), mean-class Top-5 (mCA@5), and macro-F1 (MF1), which provide a more balanced assessment than overall accuracy.

\noindent\textbf{Prompt Design.} We compare two prompt variants for NLL-based inference: with and without the candidate label list. The results show that this choice leads to noticeable performance differences. To keep a fair comparison, we adopt the NLL prompt whose semantic content is aligned with that used in free-form generation, meaning that we try to keep those prompts the same with the least modification. Detailed results are provided in SM E.2.

\subsection{Why Vanilla MLLMs Struggle with MGR: A Two-View Analysis}
\label{Sec:Anlaysis_of_failure}

We first analyze why vanilla MLLM inference has poor performance on MGR based on two bottlenecks.

\begin{figure}[htbp]
	\centering
    \begin{subfigure}[t]{0.3\linewidth}
        \centering
        \includegraphics[width=0.95\linewidth]{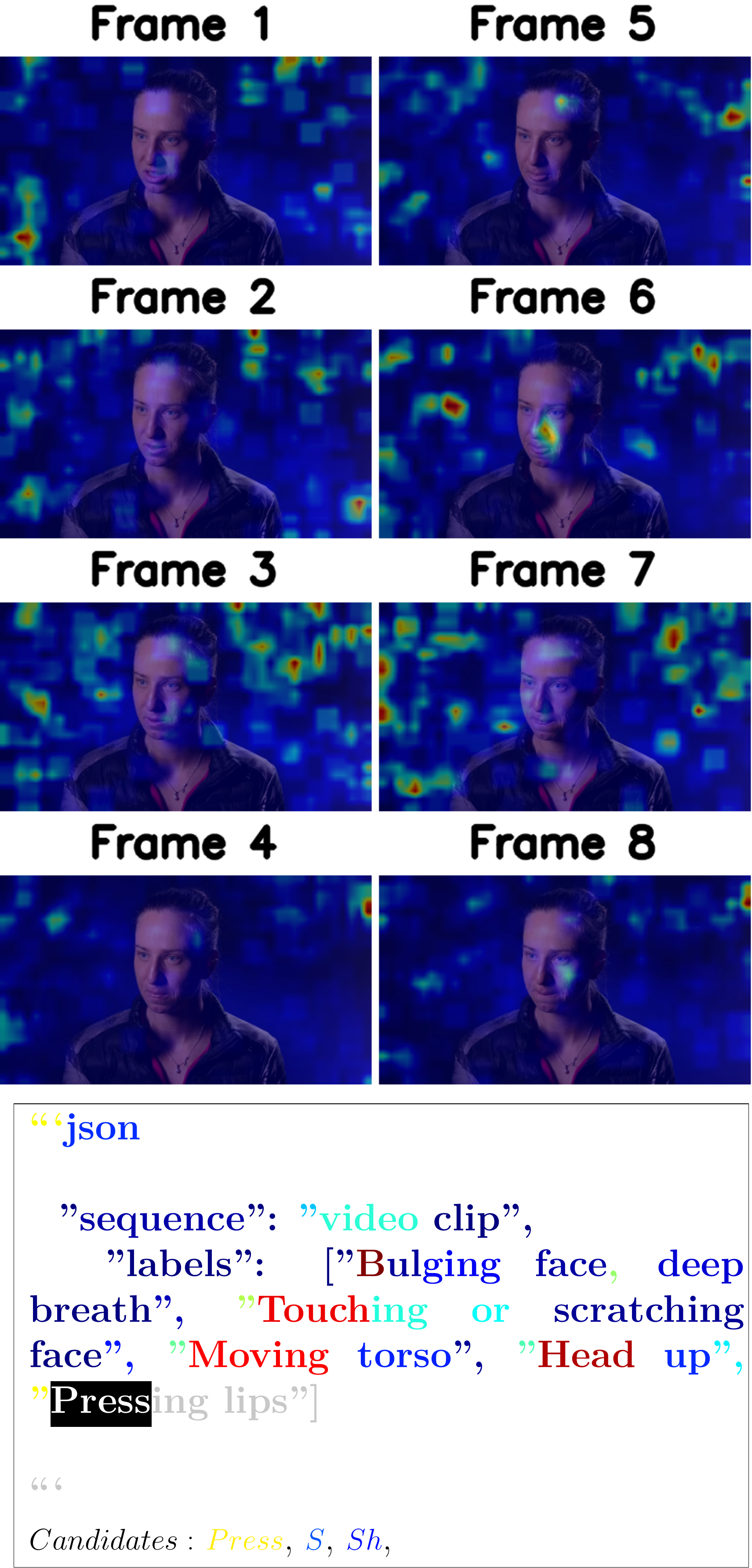}
        \caption{}
        \label{fig:0041}
    \end{subfigure}
    \begin{subfigure}[t]{0.3\linewidth}
        \centering
        \includegraphics[width=0.95\linewidth]{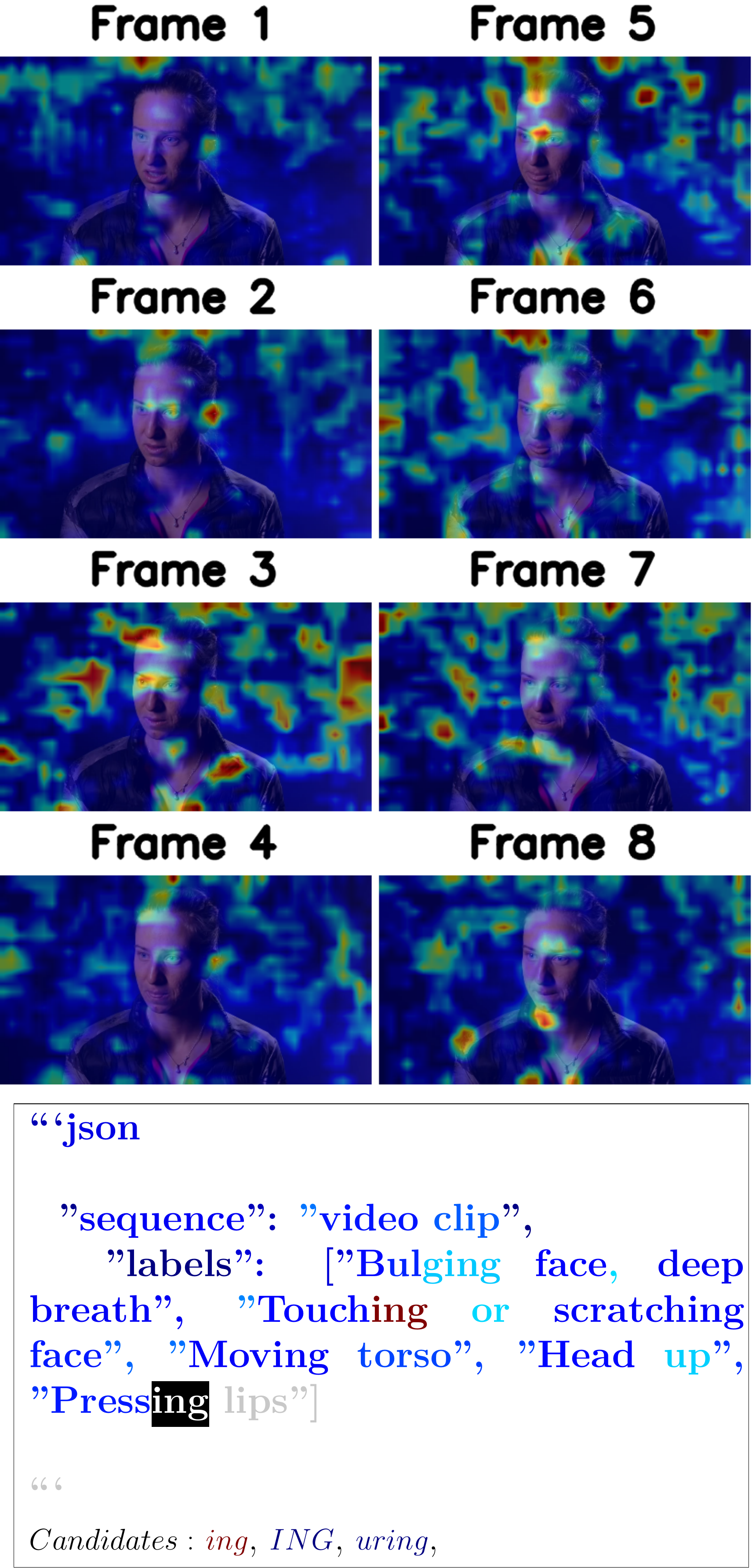}
        \caption{}
        \label{fig:0042}
    \end{subfigure}
    \begin{subfigure}[t]{0.3\linewidth}
        \centering
        \includegraphics[width=0.95\linewidth]{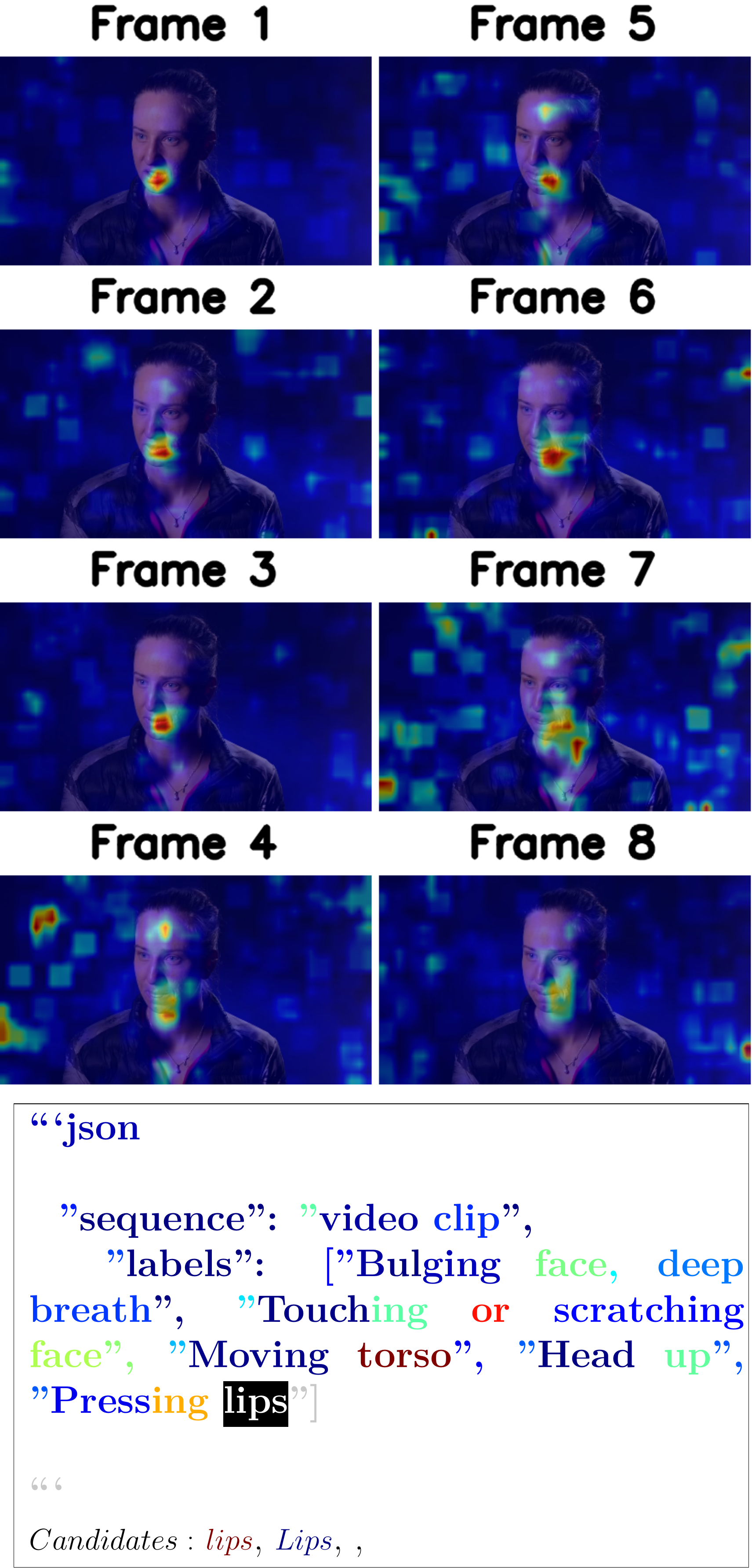}
        \caption{}
        \label{fig:0043}
    \end{subfigure}
	\caption{The TAM visualization of a failure case for the gesture \textit{Pressing lips}. The wrong prediction is \textit{Bulging face, deep breath}. We show the token-wise activation maps for three successive generated tokens during free-form generation in (a), (b), and (c), respectively. Although the model suggests the GT in Top-5, the highlighted regions are not consistently grounded on the lip area or the most discriminative motion phase, indicating that the naive single-pass MLLM fails to capture compact motion-centric local evidence.}
    \label{fig:failure_figure_tam_vis}
\end{figure}

First, under the common general prompting setting, i.e., prompt-based free-form generation, vanilla MLLMs exhibit weak grounding on discriminative local evidence. To better analyze this limitation, we employ TAM \cite{tam} for token-level visual attribution. The resulting maps suggest that, even when the model partially captures the correct semantics, it often fails to consistently attend to the truly decisive regions, such as the lips area. Such insufficient localization therefore leads to misclassification. A representative failure case visualized by TAM is shown in Fig.~\ref{fig:failure_figure_tam_vis}. In contrast, when effective local evidence is explicitly leveraged, the correct prediction can be recovered, which is provided in SM F.1.

\begin{figure}[htbp]
	\centering
    \begin{subfigure}[t]{0.95\linewidth}
        \centering
        \includegraphics[width=0.9\linewidth]{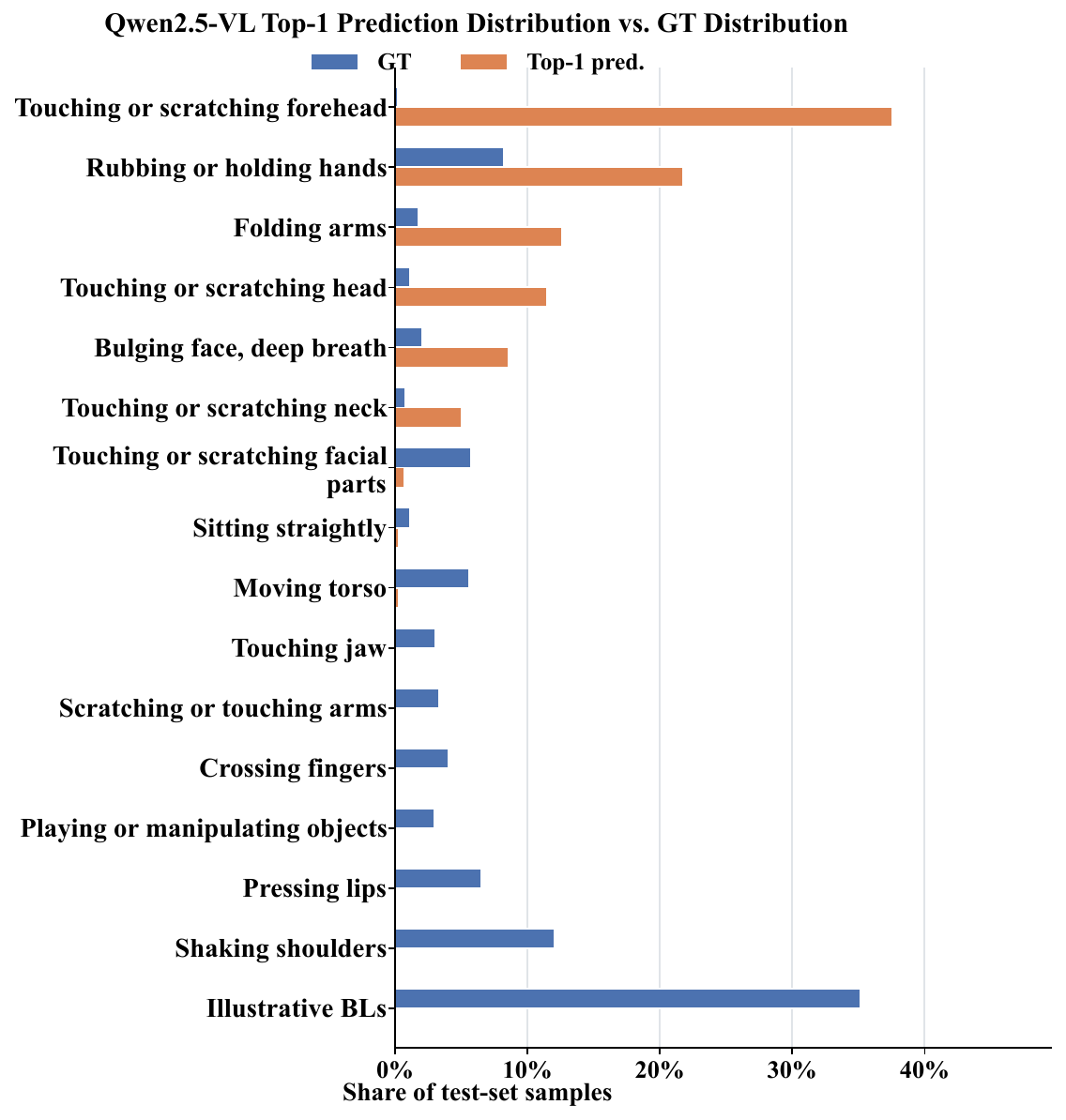}
        \caption{}
        \label{fig:failure_anlaysis_figure_pred_top1_collapse}
    \end{subfigure}
    \begin{subfigure}[t]{0.95\linewidth}
        \centering
        \includegraphics[width=0.9\linewidth]{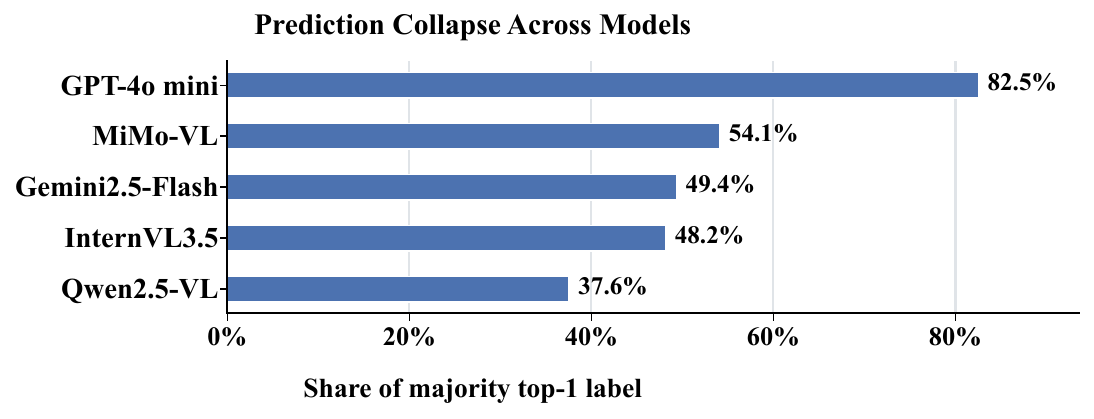}
        \caption{}
        \label{fig:failure_anlaysis_figure_cross_model}
    \end{subfigure}
	\caption{Failure analysis on iMiGUE of the prompt-based free-form generation. (a) \textbf{The comparison of distributions of Qwen2.5-VL predictions and the ground-truths (GTs) of iMiGUE}. Due to the space limitations, only the appearing frequency of labels is larger than 1\% in either GT or Top-1 prediction is listed. (b) \textbf{The comparison of the severity of the prediction-collapse across different vanilla MLLMs}.}
    \label{fig:failure_analysis_two_figures}
    \vspace{-0.3cm}
\end{figure}

Second, under free-form generation, Qwen2.5-VL shows a clear prediction-collapse pattern. As shown in Fig.~\ref{fig:failure_anlaysis_figure_pred_top1_collapse}, a large portion of test clips is mapped to a few dominant labels, leading to poor coverage of the label space and a distribution inconsistent with the ground-truths (GTs). 
This suggests that \textbf{the visual input is often insufficient to override the language priors of the model, namely the language bias}. Besides, this prediction-collapse pattern is observed across various MLLMs, as revealed by Fig.~\ref{fig:failure_anlaysis_figure_cross_model}. Moreover, the bias-calibrated NLL analysis is conducted to quantify prior-driven score distortion. Due to the space limitation, relevant results and figures are listed in SM F.2.
The NLL analysis indicates that the two calibrations introduced address overlapping but distinct subsets of errors, suggesting that language bias and appearance bias are complementary rather than redundant. These observations motivate the two key components of Zero-MELO: localized evidence acquisition and dual-reference calibration.

\subsection{Experimental Results}

\begin{table*}[t]
    \centering
    \caption{Zero-shot performance comparison on iMiGUE testset and MA-52 valid set.}

    \begin{tabular}{@{}l ccc @{\hspace{3em}} ccc@{}}
        \toprule
        \multirow{2}{*}{\textbf{Method}} & \multicolumn{3}{c}{\textbf{iMiGUE}} & \multicolumn{3}{c}{\textbf{MA-52}} \\
        \cmidrule(lr){2-4} \cmidrule(l){5-7}
        & \textbf{mCA@1} & \textbf{mCA@5} & \textbf{MF1} & \textbf{mCA@1} & \textbf{mCA@5} & \textbf{MF1} \\
        \midrule
        
        \multicolumn{7}{c}{\textit{Prompt-based Zero-shot}} \\
        \midrule
        GPT-4o mini \cite{openai_gpt4o_2024} & 0.1113 & 0.3392 & 0.0699 & 0.0576 & 0.2362 & 0.0363 \\
        Gemini2.5-Flash \cite{gemini25} & 0.1917 & 0.3689 & 0.1367 & 0.1666 & 0.3072 & 0.1217 \\
        MiMo-VL-7B \cite{mimovl} & 0.2445 & 0.5643 & 0.1506 & 0.1869 & 0.3814 & 0.1042 \\
        InternVL3.5-8B \cite{internvl35} & 0.2034 & 0.4358 & 0.0927 & 0.1796 & 0.3677 & 0.0914 \\
        Qwen2.5-VL-7B \cite{qwen25vl} & 0.1615 & 0.4101 & 0.0728 & 0.1020 & 0.3133 & 0.0680 \\
        \midrule
        
        \multicolumn{7}{c}{\textit{NLL-Based}} \\
        \midrule
        MiMo-VL-7B & 0.0325 & 0.2480 & 0.0008 & 0.0402 & 0.1521 & 0.0144 \\
        InternVL3.5-8B & 0.0329 & 0.2408 & 0.0007 & 0.0721 & 0.2387 & 0.0414 \\
        Qwen2.5-VL-7B & 0.0523 & 0.2524 & 0.0072 & 0.1178 & 0.3299 & 0.0714 \\
        \midrule
        
        \multicolumn{7}{c}{\textit{Test-Time Method}$^{*}$} \\
        \midrule
        ZoomEye $^{**}$ \cite{zoomeye} & 0.0760 &  0.2340 & 0.0326 & 0.1185 & 0.3508 & 0.0723 \\
        VCD \cite{VCD} & 0.2077 & 0.4717 & 0.1045 & 0.1516
 & 0.3364 & 0.0888 \\
        Zero-MELO & \textbf{0.2684} & \textbf{0.5775} & 0.1325 & \textbf{0.2210} & \textbf{0.4359} & \textbf{0.1221} \\
        \bottomrule
    \end{tabular}
    \vspace{2pt}
    \parbox{0.95\textwidth}{\footnotesize
    \raggedright
    $^{*}$ All test-time methods in this table use Qwen2.5-VL as the common backbone for fair comparison. \\
    $^{**}$ ZoomEye was originally proposed for image inputs. For this comparison, we adapt it to process video inputs and conduct it in the NLL-based setting.
    }
    \label{table:main_results}
    \vspace{-0.1cm}
\end{table*}

\paragraph{\textbf{Comparison with SOTA Models}} 
We first evaluate zero-shot MGR performance on both iMiGUE and MA-52. As shown in Table~\ref{table:main_results}, Zero-MELO is compared with SOTA vanilla MLLMs and representative test-time baselines, including ZoomEye and VCD. The results demonstrate that 
Zero-MELO achieves the best mCA@1 and mCA@5, effectively mitigating the limitations of global-view MLLM inference in zero-shot MGR.

\paragraph{\textbf{NLL-Based vs. Prompt-based Generation}}
As discussed in Sec.~\ref{Sec:Methodology} regarding the technical path choice of MLLMs, we compare the results of NLL-based zero-shot and prompt-based free-form zero-shot in Table~\ref{table:main_results}. 
For most compared models, NLL-based inference is weaker than prompt-based free-form generation under the current prompt design, due to the amplified prediction-collapse problem.
For instance, MiMo-VL ranks \textit{Buckle button, pulling shirt collar, adjusting tie} as the Top-1 prediction for nearly all samples despite minor score variation across samples. 
In contrast, applying blank calibration substantially improves the NLL-based results and significantly narrows the gap to free-form generation, which supports the effectiveness of the proposed calibration. For instance, InternVL3.5 improves from $3.29\%$ to $17.68\%$, yielding an absolute gain of $14.39$ percentage points. More detailed analyses are discussed in SM G. 

Moreover, we also observe a dataset-related reversal in Table~\ref{table:main_results}. While prompt-based baselines are consistently stronger on iMiGUE, NLL-based vanilla MLLMs are consistently stronger on MA-52. This counterintuitive gap suggests that the behavior of raw NLL ranking is not determined solely by visual difficulty, but is also strongly affected by dataset-dependent prior structure in the label space. After calibration, this reversal becomes less consistent, indicating that calibration and evidence refinement partially reduce the bias. Detailed analyses are also provided in SM G.

\paragraph{\textbf{SFT vs Zero-shot for MLLM}}
Since zero-shot performance does not represent the upper bound of MLLM capability on MGR, we further examine how far a vanilla MLLM can be improved through supervised fine-tuning. Specifically, we fine-tune Qwen2.5-VL under two task formulations: an NLL-style label-scoring formulation and a VQA-style formulation. 
As shown in Table~\ref{table:3}, both formulations improve over the vanilla zero-shot baseline, and the VQA-style formulation performs better than the label-scoring formulation. However, both remain below Zero-MELO, suggesting that our test-time framework is more effective than straightforward SFT under the long-tailed label distribution.

\begin{table}[t]
    \centering
    \caption{Comparison between our method and the SFT method on MA-52 dataset.}
    \begin{tabular}{@{}l ccc @{\hspace{3em}} ccc@{}}
        \toprule
        \textbf{Method} & \textbf{mCA@1} & \textbf{mCA@5} & \textbf{MF1} \\
        \midrule
        VQA & 0.177 & 0.4201 & 0.0842  \\
        Label Scoring & 0.0899 & 0.2545 & 0.0623  \\
        \midrule
        Ours & \textbf{0.2210} & \textbf{0.4359} & \textbf{0.1221} \\
        \bottomrule
    \end{tabular}
    \label{table:3}
    \vspace{-0.1cm}
\end{table}



\begin{figure*}[h]
    \centering
    \includegraphics[width=\linewidth]{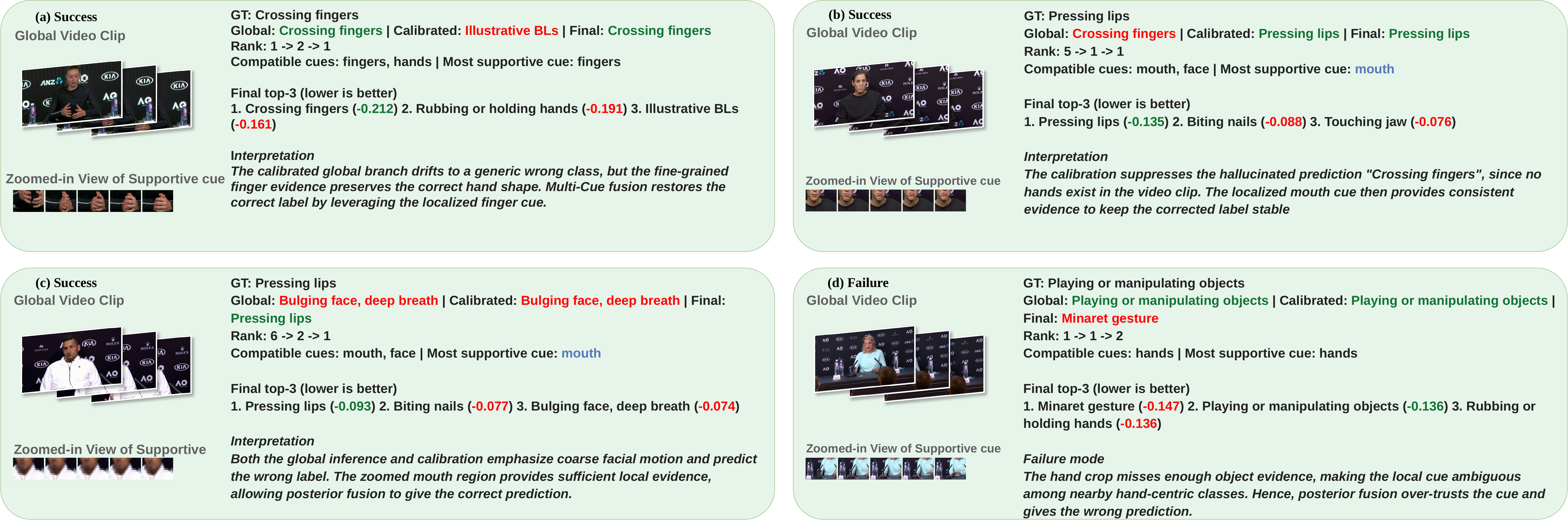}
    \caption{Case study of Zero-MELO. Each panel shows the global video clip, the zoomed-in view of the most supportive cue, and the trajectory of the prediction Top-1 across global inference, calibration, and final multi-cue fusion. The \textbf{(a)}, \textbf{(b)}, \textbf{(c)} correspond to the successful samples, while \textbf{(d)} is the failure case.}
    \label{fig:case_study}
\end{figure*}    

\subsection{Ablation Study}
\paragraph{\textbf{Different Backbone}}
Different backbones are explored with Zero-MELO. The full results are provided in SM I, 
where Zero-MELO consistently improves over the corresponding vanilla MLLMs.

\paragraph{\textbf{Different Test-Time Calibration Strategies}}
The blank and lowrank calibrations are ablated, respectively. The results are listed in SM J
since the page limitation is strict. The experiment indicates that the joint fusion of these two calibrations can contribute to better performance, aligning with Sec.~\ref{Sec:Anlaysis_of_failure} that the two calibrations proposed aim to mitigate different biases and are complementary.


\paragraph{\textbf{Component Validity}}
Each component of Zero-MELO is ablated in Table~\ref{table:component_validity}. Note that the vanilla baseline in this table is not numerically identical to that in Table~\ref{table:main_results}, due to the structural difference of default visual input and the minor prompt alternatives. Check SM K 
for details.

Note that purely adding the tree search module leaves mCA@1 almost unchanged. This is reasonable since the language prior bias is dominant and thus changes of visual input do not affect Top-1 prediction, while the additional local evidence mainly moves the ground-truth label upward in the ranking, improving mCA@5 (more useful information). Considering such a strong language prior, we use blank calibration to first mitigate the bias to compare the benefits of the search module fairly. As such, the more sufficient local evidence acquired via the search algorithm contributes to the higher mCA@1. Notably and encouragingly, simply applying blank calibration allows significant improvement of mCA@1 from $0.0357$ to $0.2040$, supporting the existence of the dominant language prior. The effectiveness of the search module under blank calibration also suggests that the insufficient local evidence and severe biases are not identical but complementary reasons for the failures of MLLMs in MGR. Besides, aggregate calibration further improves performance, which supports the analysis in Sec.~\ref{Sec:Anlaysis_of_failure}. Finally, multi-cue fusion yields consistent gains, and our strategy achieves the best performance. Details of the experiment are illustrated in SM~K.


\begin{table}[t]
    \centering
    \caption{Ablation study of different components of Zero-MELO on iMiGUE dataset. Vanilla (ablation format) uses the unified paired-input format adopted by all component ablations, differing from Table~\ref{table:main_results}. TS, Cali, and F denote tree search, calibration, and fusion; Max and Average denote lowest-NLL cue selection and uniform cue averaging.
    }
    \begin{tabular}{@{}lccc@{}}
        \toprule
        \textbf{Method} & \textbf{mCA@1} & \textbf{mCA@5} & \textbf{MF1} \\
        \midrule
        Vanilla (ablation format)  & 0.0357 & 0.2084 & 0.0094\\
        + TS & 0.0358 & 0.2629 & 0.0100 \\
        + Cali (Blank)  & 0.2040 & 0.4943 & 0.0855\\
        + Cali (Aggregate) & 0.2363 & 0.5436 & 0.1097\\
        + TS + Cali (Aggregate, Single-Cue) & 0.2244 & 0.5342 & 0.1204 \\
        + TS + Cali (Blank)  & 0.2131 & 0.5078 & 0.0924\\
        + TS + Cali + F (Max)  & 0.2287 & 0.5365 & 0.1227\\
        + TS + Cali + F (Average)  & 0.2508 & 0.5621 & 0.1278\\
        + TS + Cali + F (\textbf{Ours}) & \textbf{0.2684} & \textbf{0.5775} & \textbf{0.1325}\\
        \bottomrule
    \end{tabular}
    \label{table:component_validity}
    \vspace{-0.15cm}
\end{table}


\subsection{Case Analysis}
Fig.~\ref{fig:case_study}(a)--(c) illustrate three complementary behaviors: localized evidence can recover fine-grained hand or mouth cues, while calibration suppresses prior-driven global errors. Fig.~\ref{fig:case_study}(d) shows the main failure mode: an overly tight hand crop removes object evidence, which leads to a nearby hand-centric class.


\subsection{Limitation}

Although Zero-MELO compares favorably with existing MLLM-based methods, it still has two main limitations. First, Zero-MELO remains sensitive to cue localization: missed or overly tight crops can discard discriminative object or context evidence. 
Second, Zero-MELO introduces substantial test-time overhead due to multi-cue search and repeated label scoring, despite practical caching and batching. It requires 399.82 seconds and 585 MLLM calls per iMiGUE sample on an NVIDIA A100 40GB GPU. Detailed comparisons are provided in SM~M. 
\section{Conclusion}
In this paper, we first revisit the failure patterns of MLLMs in MGR. We observe that MLLMs suffer from two bottlenecks: insufficient localized evidence and severe score biases driven by language and motion-agnostic appearances. Accordingly, we present Zero-MELO. Extensive experiments validate the effectiveness of Zero-MELO over different backbones and datasets. Our results suggest that the limitations of current MLLMs on MGR are twofold: a substantial portion comes from insufficient elicitation under standard inference, while the remaining hard cases indicate that current models still have limited capability on extremely fine-grained motion-centric understanding.
\begin{acks}



This work was supported by the Research Council of Finland Research Fellow project (grant 371019), flagship Profi 7 (grant 352788), University of Oulu, and Infotech Oulu. We also wish to acknowledge CSC– IT Center for Science, Finland, for computational resources.
\end{acks}




\bibliographystyle{ACM-Reference-Format}
\bibliography{sample-base}

\appendix
\section{Misaligned Cases in Prompt-Based Free-Form Generation of Close-Source MLLMs}
\label{appendix:misaligned_failure_predicitions}

This section provides additional examples showing that, under prompt-based free-form generation, closed-source MLLMs may produce misaligned labels, i.e., the labels whose names do not exactly match the predefined label inventory. We illustrate this phenomenon using GPT-4o mini and Gemini2.5-Flash on both iMiGUE and MA-52.

We count a clip as mismatched if at least one returned top-$k$ label is not an exact-match dataset label, without applying alias normalization or fuzzy repair. Even when the prompt explicitly requires the model to output exact strings from the provided label set, such misaligned outputs still occur.

Table~\ref{tab:misaligned_stats} summarizes the mismatch statistics on the two datasets. On iMiGUE, GPT-4o mini and Gemini2.5-Flash produce exact-string mismatches on 36 and 280 clips, respectively. On MA-52, the corresponding numbers are 441 and 195. These statistics are reported only to show that the phenomenon is systematic rather than anecdotal. Table~\ref{tab:misaligned_examples} lists several representative examples.

\begin{table}[t]
\centering
\small
\caption{Simple statistics of exact-name mismatches in free-form generation. A clip is counted if at least one returned top-$k$ label is not an exact-string-matched label.}
\label{tab:misaligned_stats}
\begin{tabular}{l l r r}
\toprule
Dataset & Model & Total clips & Clips with mismatched labels \\
\midrule
iMiGUE & GPT-4o mini & 4534 & 36 (0.79\%) \\
iMiGUE & Gemini2.5-Flash & 4534 & 280 (6.18\%) \\
MA-52 & GPT-4o mini & 5586 & 441 (7.89\%)\\
MA-52 & Gemini2.5-Flash & 5586 & 195 (3.49\%)\\
\bottomrule
\end{tabular}
\end{table}

\begin{table}[t]
\centering
\small
\caption{Representative misaligned labels observed in prompt-based free-form generation.}
\label{tab:misaligned_examples}
\begin{tabular}{l l p{0.54\linewidth}}
\toprule
Dataset & Model & Representative misaligned labels \\
\midrule
iMiGUE & GPT-4o mini &
Touching or scratching jaw; Touching forehead; Touching or adjusting hair \\
iMiGUE & Gemini2.5-Flash &
Holding back arms; Adjusting tie; Touch jaw; Touchng jaw \\
MA-52 & GPT-4o mini &
Bowling head; Scratching or touching head; Raising up; Touching face \\
MA-52 & Gemini2.5-Flash &
Scratching face; Touching face; Touching chest; Tipping head \\
\bottomrule
\end{tabular}
\end{table}

\section{Implementation Details of Cue-Guided Tree Search} 
\label{appendix:implementation_details_of_zoom_search}

This section supplements Sec.~3.3 with the concrete implementation used to instantiate the anchor-wise search operator: \(\mathrm{search}(g_a,c_i)\), the retained proposal set: \(\hat{\mathcal{B}}_{i,a}\), and the aggregation operator: \(\mathrm{Agg}(\cdot)\). Since the tree search component is inspired by the generic single-image tree-exploration backbone, we do not restate its generic node bookkeeping or image-tree traversal here. Instead, we focus only on the implementation choices that are specific to our video-based MGR setting.

\paragraph{\textbf{Anchor-Wise Search on Local Temporal Windows.}}
Since the target is a video clip containing people's movements, search is performed on a small anchor set \(\mathcal{A}\) on the sampled timeline. For each anchor \(a\in\mathcal{A}\), the search tree itself is built on the corresponding anchor frame \(g_a\), i.e., tree nodes remain spatial boxes on a single image. However, each MLLM confidence probe is evaluated on a short anchor-centered local clip rather than on a single snapshot. This design preserves short-term motion context for transient micro-actions while keeping the number of MLLM queries bounded. Inspired by the idea that the onset, apex, and offset frames can describe a micro-expression, we use three anchors in our experiments.

\paragraph{\textbf{Semantic Misalignment Mitigation.}}
The spatial search space is discretized by a hierarchical patch tree constructed on each anchor frame. A limitation of rigid, non-overlapping patches is the semantic misalignment: small cue regions may straddle partition boundaries and be partially cropped in every child node, reducing cue detectability.
To alleviate this, an \emph{overlapping partition} is adopted when splitting a parent box of width $W$ and height $H$ into $k_w\times k_h$ children. Let the parent box parametrized as $B=[\xi_0,\eta_0,W,H]$, where $(\xi_{0}, \eta_{0})$ denotes the top-left corner of the paraent box, and $W$ and $H$ denote its width and height, respectively. Then, it is split into a $k_w\times k_h$ grid using edge coordinates,
\begin{equation}
\begin{aligned}
    &\xi_u = \xi_0 + \mathrm{round}\!\left(\frac{uW}{k_w}\right),~u=0,\dots,k_w,\\
&\eta_v = \eta_0 + \mathrm{round}\!\left(\frac{vH}{k_h}\right),~v=0,\dots,k_h,
\end{aligned}
\end{equation}
with $\xi_{k_w}=\xi_0+W$ and $\eta_{k_h}=\eta_0+H$.
For each cell $(u,v)$, denote its base width and height as $w_b=\xi_{u+1}-\xi_u$ and $h_b=\eta_{v+1}-\eta_v$.
With overlap ratio $\rho\in[0,1)$, the cell is expanded symmetrically and clipped such that it lies within the parent box:
\begin{equation}
\begin{aligned}
    &\tilde{\xi}_1=\max (\xi_0, \xi_u-\frac{\rho}{2}w_b),~~\tilde{\xi}_2=\min(\xi_0+W, \xi_{u+1}+\frac{\rho}{2}w_b),\\
&\tilde{\eta}_1=\max(\eta_0, \eta_v-\frac{\rho}{2}h_b),~~\tilde{\eta}_2=\min(\eta_0+H, \eta_{v+1}+\frac{\rho}{2}h_b).
\end{aligned}
\end{equation}
In this case, the resulting overlapping child box is 
\begin{equation}
    \tilde{B}_{u,v}=[\tilde{\xi}_1,\tilde{\eta}_1,\tilde{\xi}_2-\tilde{\xi}_1,\tilde{\eta}_2-\tilde{\eta}_1] \subseteq B.
\end{equation}
The terminal tree patch size is set smaller than the final resized MLLM input, which yields finer localization while leaving the downstream visual encoder unchanged.

\paragraph{\textbf{Node Ranking, Stopping, and Threshold Relaxation.}}
Tree expansion is conducted in a best-first manner. Each queued node is assigned a fast confidence used only for search priority. At shallow depths, the priority score is computed based on a \emph{latent prompt} that asks whether further zooming-in would likely help. At deeper depths, the priority switches to an \emph{existence prompt} that asks whether the cue is already present in the current region. Denoting the node depth by \(d(n)\), the priority score is
\begin{equation}
s(n)=
\begin{cases}
s_{\mathrm{lat}}(n), & d(n)\le d_0,\\
s_{\mathrm{ex}}(n), & d(n)>d_0,
\end{cases}
\end{equation}
where both \(s_{\mathrm{lat}}(n)\) and \(s_{\mathrm{ex}}(n)\) are obtained from the binary \texttt{Yes}/\texttt{No} logits of the MLLM. Nodes are re-ranked by \(s(n)\) after each expansion. Stopping is decided by a cue-specific adequacy query, i.e., \emph{cue question} applied to the current zoomed-in view.  At inference time, the \textit{latent prompt}, \textit{existence prompt}, or \textit{cue question} is combined with a wrapper prompt before being fed to the model.

\paragraph{\textbf{Anchor-Wise Proposal Retention and Robust Aggregation.}}
For each anchor \(a\), the search returns a candidate set \(\mathcal{B}_{i,a}\). After removing duplicate boxes, we retain only a small subset \(\hat{\mathcal{B}}_{i,a}\subseteq\mathcal{B}_{i,a}\) ranked by stopping confidence, with depth used as a tie-breaker. The retained proposals are then aggregated across anchors. Let
\begin{equation}
\mathcal{U}_i=\bigcup_{a\in\mathcal{A}}\hat{\mathcal{B}}_{i,a}.
\end{equation}
Rather than directly unioning all boxes, we first cluster \(\mathcal{U}_i\) by pairwise IoU to suppress anchor-specific outliers. If a dominant IoU-consistent cluster exists, only that cluster is retained; otherwise all available boxes are kept. The surviving boxes are then unioned, padded by a small ratio relative to the union box size, and clipped to the image boundary. The resulting box is used as the final cue region \(b_i\). If no reliable proposal is obtained, or if the aggregated box becomes excessively large relative to the full frame, we conservatively fall back to the root region. This yields a temporally stable tube-like crop while remaining robust to occasional anchor failures.

\paragraph{\textbf{Rendering of the Local Evidence of Each Cue.}}
Once \(b_i\) is obtained, the same spatial box is applied to every sampled frame:
\begin{equation}
z_i=\mathrm{CropResize}(g,b_i).
\end{equation}

\paragraph{\textbf{Batched Yes/No Evaluation and caching.}}
All confidence computations in Step~1 are binary \texttt{Yes}/\texttt{No} probes. Nodes sharing the same prompt type and view format are grouped into batches before being sent to the MLLM. Only the logits of the \texttt{Yes} and \texttt{No} tokens are used to compute the confidence score.

In addition, decoded sampled frames, anchor-centered local windows, rendered node views, and processor-side visual tensors are cached and reused across repeated queries. When multiple candidate labels invoke the same cue, the corresponding search is executed only once, and the resulting cue-conditioned view is reused in later all-label scoring. These optimizations reduce repeated MLLM forwards without changing the search objective.

\section{Detailed Derivation of Test-Time Calibration} 
\label{appendix:detailed_derivation_calibration}
In this section, the detailed derivation of test-time calibration in Sec.~3.4 is discussed. Based on Eq.~(5) in the main paper, let $v(y)=\bigl(v_1(y),\dots,v_{|v(y)|}(y)\bigr)$ denote the tokenized label string for the candidate label $y$, and let
$
v_{<j}(y)=\bigl(v_1(y),\dots,v_{j-1}(y)\bigr)
$.
Under teacher forcing, the conditional likelihood factorizes as
\begin{equation}
p_{\mathcal{M}}\!\left(v(y)\mid x,t\right)
=
\prod_{j=1}^{|v(y)|}
p_{\mathcal{M}}\!\left(v_j(y)\mid v_{<j}(y),x,t\right),
\end{equation}
and thus
\begin{equation}
\log p_{\mathcal{M}}\!\left(v(y)\mid x,t\right)
=
\sum_{j=1}^{|v(y)|}
\log p_{\mathcal{M}}\!\left(v_j(y)\mid v_{<j}(y),x,t\right).
\end{equation}

In our implementation, only the answer tokens in $v(y)$ are scored, while the prompt prefix is masked out. Therefore, the label-wise score is the mean token-level negative log-likelihood
\begin{equation}
\ell_x(y)
=
-\frac{1}{|v(y)|}
\sum_{j=1}^{|v(y)|}
\log p_{\mathcal{M}}\!\left(v_j(y)\mid v_{<j}(y),x,t\right).
\label{eq:appendix_nll_impl}
\end{equation}
The prediction is then obtained by
\begin{equation}
y^*=\arg\min_{y\in Y}\ell_x(y).
\end{equation}
In practice, we compute token-level cross-entropy, which is equivalent to NLL under teacher forcing. Therefore, the label-wise score is
\begin{equation}
\begin{aligned}
\ell_x(y)
&=
\frac{1}{|v(y)|}
\sum_{j=1}^{|v(y)|}
\mathrm{CE}\!\left(
\delta_{v_j(y)},
p_{\mathcal{M}}(\cdot \mid v_{<j}(y),x,t)
\right) \\
&=
-\frac{1}{|v(y)|}
\sum_{j=1}^{|v(y)|}
\log p_{\mathcal{M}}\!\left(
v_j(y)\mid v_{<j}(y),x,t
\right),
\end{aligned}
\label{eq:appendix_ce_nll_equiv}
\end{equation}
where $\delta_{v_j(y)}$ denotes the one-hot target distribution of the $j$-th answer token.

To diagnose the bias, we leverage the introduced $x_b$ and $x_r$ to obtain the corresponding scores $s_{x_b}(y)$ and $s_{x_r}(y)$:
\begin{equation}
s_{x_b}(y)\triangleq -\ell_{x_b}(y), \qquad
s_{x_r}(y)\triangleq -\ell_{x_r}(y).
\end{equation}
The inference based on $x_b$ reflects language prior, while the inference based on $x_r$ captures motion-agnostic appearance prior beyond language prior via
\begin{equation}
\Delta_r(y) = s_{x_r}(y) - s_{x_b}(y).
\end{equation}
In this sense, blank and lowrank calibrations expose two partially separable prior-driven bias components in label scoring.

Let $y_t$ and $y_w$ denote the ground-truth and baseline prediction, respectively. We quantify bias as
\begin{equation}
A_{\mathrm{blank}}(x) = s_{x_b}(y_w) - s_{x_b}(y_t), \quad
A_{\mathrm{lowrank}}(x) = \Delta_r(y_w) - \Delta_r(y_t),
\end{equation}
revealing language and appearance biases, respectively. In other words, a positive $A_{\mathrm{blank}}(x)$ indicates that the blank reference favors the wrong label over the true label. A positive $A_{\mathrm{lowrank}}(x)$ indicates that, relative to the blank reference, the lowrank reference provides more support to the wrong label than to the true label. These quantities are used in SM~\ref{appendix:analysis_nll_based} to analyze the separability of the two prior components.

Based on this diagnosis, we introduce the following test-time calibration:
\begin{equation}
\tilde{s}_{x_b}(y) = s_x(y) - s_{x_b}(y), \quad
\tilde{s}_{x_r}(y) = s_x(y) - s_{x_r}(y).
\end{equation}

The prediction under each calibration is given by
\begin{equation}
y_b^* = \arg\max_y \tilde{s}_{x_b}(y), \quad
y_r^* = \arg\max_y \tilde{s}_{x_r}(y).
\end{equation}
Equivalently, in NLL form, 
\begin{equation}
\tilde{\ell}_{x_b}(y)\triangleq \ell_{x}(y)-\ell_{x_{b}}(y),
\qquad
\tilde{\ell}_{x_r}(y)\triangleq \ell_{x}(y)-\ell_{x_{r}}(y).
\end{equation}
If a label remains highly preferred even under the blank reference, it is likely dominated by language bias and is penalized by $\tilde{s}_{x_b}(y)$. Likewise, if a label is strongly supported by static appearance alone, it is penalized by $\tilde{s}_{x_r}(y)$. Therefore, the two calibrations correct different bias tendencies in the label scoring space.

\section{Detailed Derivation of Multi-Cue Fusion} 
\label{appendix:detailed_derivation_multi_cue_fusion}
Consider the following approximate conditional factorization:
\begin{equation}
    p(y\mid x_{1},\dots x_{n})\approx \prod_{i=1}^n p(y\mid x_{i}).
    \label{eq:poe}
\end{equation}
A Naive Bayes approximation is adopted, meaning that the local views are treated as independent evidence sources conditioned on $(y,g)$:
\begin{equation}
    p(z_{1:n}\mid y,g)\approx \prod_{i=1}^n p(z_i\mid y,g).
    \label{eq:cond_indep}
\end{equation}
By Bayes' rule, the posterior can be factorized as
\begin{equation}
p(y\mid g, z_{1:n}) \propto p(y\mid g)\prod_{i=1}^n p(z_i\mid y,g).
\label{eq:poe_start}
\end{equation}
Based on Bayes' theorem, 
\begin{equation}
    p(y\mid g, z_{i})=\frac{p(z_{i}\mid y,g)p(y\mid g)}{p(z_{i}\mid g)}.
    \label{eq:bayes_theorem}
\end{equation}
Substituting Eq.~\eqref{eq:bayes_theorem} into Eq.~\eqref{eq:poe_start}, and 
absorbing the terms independent of $y$ (e.g., $\prod_i p(z_i\mid g)$) into the proportionality constant, we obtain:
\begin{equation}
    p(y\mid g, z_{1:n}) \propto p(y\mid g)\prod_{i=1}^{n}\frac{p(y\mid g, z_{i})}{p(y\mid g)}=p(y\mid g)^{1-n}\prod_{i=1}^{n}p(y\mid g,z_{i}).
    \label{eq:final}
\end{equation}

\paragraph{\textbf{Implementation in Calibrated NLL Space}}

In particular, $\bar{\ell}_{x}(y)$ denotes $\bar{\ell}_{g}(y)$ for global view input (whole frame) and $\bar{\ell}_{i}(y)$ for a zoomed-view input (cropped local evidence).
\begin{equation}
\bar{\ell}_{g}(y)
=
-\tau_d \log \!\left(
\pi_g \exp\!\left(-\frac{\tilde{\ell}_{g_b}(y)}{\tau_d}\right)
+
(1-\pi_g)\exp\!\left(-\frac{\tilde{\ell}_{g_r}(y)}{\tau_d}\right)
\right),
\label{eq:global_dual_energy}
\end{equation}
and
\begin{equation}
\bar{\ell}_{i}(y)
=
-\tau_d \log \!\left(
\pi_c \exp\!\left(-\frac{\tilde{\ell}_{i,b}(y)}{\tau_d}\right)
+
(1-\pi_c)\exp\!\left(-\frac{\tilde{\ell}_{i,r}(y)}{\tau_d}\right)
\right).
\label{eq:local_dual_energy}
\end{equation}

Hence, $\bar{\ell}_{x}(y)$ is a temperature-smoothed lower envelope of the two bias calibration energies, which will be used in the subsequent fusion.

Let
\begin{equation}
\bar{p}_{x}(y)
\triangleq
\frac{\exp(-\bar{\ell}_{x}(y))}
{\sum_{y'\in\mathcal{Y}}\exp(-\bar{\ell}_{x}(y'))}
\label{eq:branch_surrogate_posterior}
\end{equation}
be the surrogate posterior induced by $\bar{\ell}_{x}(y)$.
Substituting Eq.~\eqref{eq:branch_surrogate_posterior} into the factorization in
Eq.~\eqref{eq:final} yields, up to a Naive-Bayesian (NB) label-independent constant,
\begin{equation}
\ell_{\mathrm{NB}}(y)
\doteq
\sum_{i=1}^{n}\bar{\ell}_{i}(y) - (n-1)\bar{\ell}_{g}(y)
=
\bar{\ell}_{g}(y)
+
\sum_{i=1}^{n}\bigl(\bar{\ell}_{i}(y)-\bar{\ell}_{g}(y)\bigr),
\label{eq:raw_residual_energy}
\end{equation}
which shows that the local evidence of each cue contributes a residual correction relative to the shared global branch.

\paragraph{\textbf{Label-Conditioned Normalized Fusion}}
Eq.~\eqref{eq:raw_residual_energy} is not yet suitable as the final rule, because not every cue is semantically relevant to every label, and direct accumulation scales with both the number and the correlation of compatible cues. To account for this, let $c_i$ denote the semantic type of cue $z_i$, and let $\mathcal{C}(y)$ be a fixed label-conditioned compatibility set. The set of cue indices is defined based on anatomy:
\begin{equation}
\mathcal{I}(y)
\triangleq
\left\{
i\in\{1,\dots,n\}
\;:\;
c_i\in\mathcal{C}(y)
\ \text{and}\
\bar{\ell}_{i}(y)<\infty
\right\}.
\label{eq:compatible_index_set}
\end{equation}


For each cue branch, a non-negative family coefficient is assigned:
\begin{equation}
w_i(y)\triangleq \lambda_{\phi(c_i)},
\label{eq:cue_weight_main}
\end{equation}
where $\phi(c_i)$ maps the cue type to its family, and
$\lambda_{\phi(c_i)}\ge 0$ is the corresponding family coefficient.

With $\mathcal{I}(y)$ and $w_i(y)$ defined above, the accumulation implied by
Eq.~\eqref{eq:raw_residual_energy} is adapted to the label-dependent setting as
\begin{equation}
r(y)
\triangleq
\sum_{i\in\mathcal{I}(y)}
w_i(y)\bigl(\bar{\ell}_{i}(y)-\bar{\ell}_{g}(y)\bigr).
\label{eq:weighted_residual_term}
\end{equation}
Here, $r(y)$ collects the additional evidence contributed by the compatible cue
branches relative to the global branch.

Since the number of compatible cues may vary across labels, directly using $r(y)$
would make the scale of the correction depend on the amount of local
evidence. We therefore normalize $r(y)$ by the total active cue weight and
define
\begin{equation}
\ell_{\mathrm{fuse}}(y)
\triangleq
\bar{\ell}_{g}(y)
+
\frac{
r(y)
}{
1+\sum_{i\in\mathcal{I}(y)} w_i(y)
}.
\label{eq:normalized_fused_energy}
\end{equation}
Equivalently,
\begin{equation}
\ell_{\mathrm{fuse}}(y)
=
\frac{
\bar{\ell}_{g}(y)
+
\sum_{i\in\mathcal{I}(y)} w_i(y)\bar{\ell}_{i}(y)
}{
1+\sum_{i\in\mathcal{I}(y)} w_i(y)
}.
\label{eq:normalized_fused_average}
\end{equation}

Therefore, each candidate label $y\in\mathcal{Y}$ is assigned a fused scalar score $\ell_{\mathrm{fuse}}(y)$, where the fusion follows a shared normalized residual form but activates only the cues compatible with $y$. The final prediction is obtained by ranking the resulting label-wise fused energies over the full label set:
\begin{equation}
y_{\mathrm{final}}=\arg\min_{y\in\mathcal{Y}} \ell_{\mathrm{fuse}}(y).
\label{eq:final_label_prediction}
\end{equation}
Equivalently, the fused energies induce the surrogate posterior
\begin{equation}
\hat{p}_{\mathrm{fuse}}(y)=\frac{\exp(-\ell_{\mathrm{fuse}}(y))}
{\sum_{y'\in\mathcal{Y}}\exp(-\ell_{\mathrm{fuse}}(y'))},
\label{eq:final_label_distribution}
\end{equation}
so that
\begin{equation}
y_{\mathrm{final}}
=\arg\max_{y\in\mathcal{Y}} \hat{p}_{\mathrm{fuse}}(y)
=\arg\min_{y\in\mathcal{Y}} \ell_{\mathrm{fuse}}(y).
\label{eq:final_decision_equivalence}
\end{equation}
Thus, the final decision is made by a global ranking of the label-wise fused energies, while the local evidence entering each score remains label-conditioned through the compatibility set $\mathcal{I}(y)$.
\paragraph{\textbf{Implementation of Verbalizer Marginalization.}}
To better separate semantically close labels, such as \textit{Scratching or touching neck} and \textit{Scratching or touching chest}, we use two verbalizers for each candidate label $y$: the raw label string $v^{\mathrm{lab}}(y)$ and a short description $v^{\mathrm{desc}}(y)$. For any branch input $x$ in Sec.~3.4, including the global branch $g$ and a cue-conditioned branch $x_i=(g,z_i)$, we compute two teacher-forced scores of $g$ and $x_i$ by Eq.~\ref{eq:appendix_ce_nll_equiv}, denoted by $\ell_x^{\mathrm{lab}}(y)$ and $\ell_x^{\mathrm{desc}}(y)$, respectively.

We then define a single branch-wise label energy by marginalizing over the two verbalizers:
\begin{equation}
\ell_x^{\mathrm{vm}}(y)
=
-\log\!\left(
\alpha_y \exp\!\left(-\ell_x^{\mathrm{lab}}(y)\right)
+
(1-\alpha_y)\exp\!\left(-\ell_x^{\mathrm{desc}}(y)\right)
\right),
\label{eq:vm_two}
\end{equation}
where the coefficient $\alpha_y\in(0,1)$ controls the relative contribution of the raw label verbalizer and the description verbalizer.

This marginalized score is used in place of $\ell_x(y)$ in the calibration stage. Accordingly,
\begin{equation}
\tilde{\ell}_{x_b}^{\mathrm{vm}}(y)
=
\ell_x^{\mathrm{vm}}(y)-\ell_{x_b}^{\mathrm{vm}}(y),
\qquad
\tilde{\ell}_{x_r}^{\mathrm{vm}}(y)
=
\ell_x^{\mathrm{vm}}(y)-\ell_{x_r}^{\mathrm{vm}}(y),
\label{eq:vm_calib}
\end{equation}
and Eq.~(9) in the main paper is applied in the same way to obtain the calibrated NLL $\bar{\ell}_x^{\mathrm{vm}}(y)$.

Importantly, this implementation does not change the mathematical form of the multi-cue fusion in Sec.~3.5. The fusion derivation only requires each branch to provide a calibrated label-wise energy. Therefore, after replacing $\bar{\ell}_g(y)$ and $\bar{\ell}_i(y)$ by their marginalized counterparts $\bar{\ell}_g^{\mathrm{vm}}(y)$ and $\bar{\ell}_i^{\mathrm{vm}}(y)$, Eq.~(14) in the main paper remains valid:
\begin{equation}
\ell_{\mathrm{fuse}}^{\mathrm{vm}}(y)
=
\frac{
\bar{\ell}_g^{\mathrm{vm}}(y)
+
\sum_{i\in \mathcal{I}(y)} w_i(y)\bar{\ell}_i^{\mathrm{vm}}(y)
}{
1+\sum_{i\in \mathcal{I}(y)} w_i(y)
}.
\label{eq:vm_fuse}
\end{equation}
Thus, verbalizer marginalization only refines how each label NLL is constructed before calibration and fusion, while leaving the overall calibration and multi-cue fusion framework mathematically unchanged.

\section{Experimental Setup} 
\subsection{Experimental Configurations} 
\label{appendix:exp_config}
This section provides the experimental configurations. The complete cue sets in iMiGUE and MA-52 are listed below. We use cues based on the anatomy and the body association.
\paragraph{\textbf{iMiGUE}}
The cues used in iMiGUE are the following:
\begin{equation}
\mathcal{C}=
\left\{
\begin{aligned}
&\texttt{arms},\ \texttt{body-hand},\ \texttt{face},\ \texttt{fingers},\ \texttt{hands},\\
&\texttt{head-hand},\ \texttt{mouth},\ \texttt{neck},\ \texttt{shoulders},\ \texttt{torso}
\end{aligned}
\right\},
\end{equation}
where $\texttt{body-hand}$ and $\texttt{head-hand}$ denote the interactions between the body and hands, and between the head and hands, respectively.
For each label \(y\), the label-conditioned cue set \(\mathcal{C}(y)\) is fixed as shown in Table~\ref{table:appendix_label_conditioned_cue_set_imigue}. There is no corresponding cue for \textit{Illustrative BLs}, since this class is not defined as an MG in iMiGUE. For the remaining tree-search parameters, we set the patch size of each node to 224. The $k_{w}$ and $k_{h}$ are both set to 2, and the overlap ratio is defined as 0.25. The tree-search batch size of the tree search is fixed to be 8. The label descriptions were generated by GPT-5.4-thinking and are provided in Table~\ref{table:appendix_descriptions}. The parameters related to multi-cue fusion are listed in Table~\ref{table:appendix_parameters_multi_cue_fusion_imigue}.

\begin{table}[t]
    \centering
    \caption{Parameters of multi-cue fusion in iMiGUE.}
    \setlength{\tabcolsep}{3pt}
    \begin{tabular}{@{}l ccc@{}}
        \toprule
        \textbf{Parameter} & \textbf{Qwen2.5-VL} & \textbf{InternVL3.5} & \textbf{MiMo-VL} \\
        \midrule
        $\tau_{d}$ & 0.9 & 0.7 & 1.1 \\
        $\pi_{g}$  & 0.5 & 0.5 & 0.5 \\
        $\pi_{c}$  & 0.5 & 0.5 & 0.5 \\
        $\alpha_{y}$ & 0 & 0.85 & 0.97\\
        $w_{head}$  & 0.2 & 1.0 & 1.5 \\
        $w_{hand}$  & 1.0 & 0.8 & 1.12 \\
        $w_{torso}$  & 0.2 & 0.4 & 0.1 \\
        $w_{other}$  & 0.5 & 0.2 & 1.0 \\
        \bottomrule
    \end{tabular}
    \label{table:appendix_parameters_multi_cue_fusion_imigue}
\end{table}

\paragraph{\textbf{MA-52}}
The cues used in MA-52 are the following:
\begin{equation}
\mathcal{C}=
\left\{
\begin{aligned}
&\texttt{feet},\ \texttt{hands},\ \texttt{hands and arms},\ \texttt{head},\ \texttt{head-hand},\\
&\texttt{leg-hand},\ \texttt{legs},\ \texttt{mouth},\ \texttt{neck},\ \texttt{torso}
\end{aligned}
\right\}.
\end{equation}
The label-conditioned cue set is listed in Table~\ref{table:appendix_label_conditioned_cue_set_ma52}. For the tree search parameters, we set the patch size to 112 since the video resolution is lower than that of iMiGUE. The $k_{w}$ and $k_{h}$ are also set to 2, and the overlap ratio is defined as 0.25. The label descriptions for MA-52 are taken from its official material, as shown in Table~\ref{table:appendix_descriptions}. The parameters related to multi-cue fusion are listed in Table~\ref {table:appendix_parameters_multi_cue_fusion_ma52}.

\begin{table}[t]
    \centering
    \caption{Parameters of multi-cue fusion in MA-52.}
    \setlength{\tabcolsep}{3pt}
    \begin{tabular}{@{}l c@{}}
        \toprule
        \textbf{Parameter} & \textbf{Qwen2.5-VL} \\
        \midrule
        $\tau_{d}$ & 0.5 \\
        $\pi_{g}$  & 0.5  \\
        $\pi_{c}$  & 0.5  \\
        $\alpha_{y}$ & 0.25 \\
        $w_{head}$  & 0.7 \\
        $w_{upper-limbs}$  & 0.1 \\
        $w_{lower-limbs}$  & 0.25 \\
        $w_{body}$  & 0.1 \\
        \bottomrule
    \end{tabular}
    \label{table:appendix_parameters_multi_cue_fusion_ma52}
\end{table}

\begin{table*}[t]
\centering
\caption{Label-conditioned cue sets \(\mathcal{C}(y)\) for iMiGUE.}
\label{table:appendix_label_conditioned_cue_set_imigue}
\small
\setlength{\tabcolsep}{4pt}
\begin{tabular}{p{0.18\textwidth} p{0.28\textwidth} p{0.18\textwidth} p{0.28\textwidth}}
\toprule
\textbf{Label \(y\)} & \textbf{Label-Conditioned Cue Set \(\mathcal{C}(y)\)} & \textbf{Label \(y\)} & \textbf{Label-Conditioned Cue Set \(\mathcal{C}(y)\)} \\
\midrule
Turtle neck & \(\{\texttt{neck}\}\) & Folding arms & \(\{\texttt{body-hand}, \texttt{arms}\}\) \\
Bulging face, deep breath & \(\{\texttt{face}\}\) & Dusting off clothes & \(\{\texttt{body-hand}, \texttt{arms}\}\) \\
Touching hat & \(\{\texttt{head-hand}\}\) & Putting arms behind body & \(\{\texttt{arms}\}\) \\
Touching or scratching head & \(\{\texttt{head-hand}\}\) & Moving torso & \(\{\texttt{torso}\}\) \\
Touching or scratching forehead & \(\{\texttt{head-hand}, \texttt{face}\}\) & Sitting straightly & \(\{\texttt{torso}\}\) \\
Covering face & \(\{\texttt{head-hand}, \texttt{face}\}\) & Scratching or touching arms & \(\{\texttt{arms}, \texttt{hands}\}\) \\
Rubbing eyes & \(\{\texttt{head-hand}, \texttt{face}\}\) & Rubbing or holding hands & \(\{\texttt{hands}\}\) \\
Touching or scratching facial parts & \(\{\texttt{face}, \texttt{head-hand}\}\) & Crossing fingers & \(\{\texttt{fingers}, \texttt{hands}\}\) \\
Touching ears & \(\{\texttt{head-hand}, \texttt{face}\}\) & Minaret gesture & \(\{\texttt{fingers}, \texttt{hands}\}\) \\
Biting nails & \(\{\texttt{head-hand}, \texttt{mouth}\}\) & Playing or manipulating objects & \(\{\texttt{hands}\}\) \\
Touching jaw & \(\{\texttt{mouth}, \texttt{head-hand}, \texttt{face}\}\) & Hold back arms & \(\{\texttt{arms}\}\) \\
Touching or scratching neck & \(\{\texttt{neck}, \texttt{head-hand}\}\) & Head up & \(\{\texttt{neck}, \texttt{face}\}\) \\
Playing or adjusting hair & \(\{\texttt{head-hand}, \texttt{face}\}\) & Pressing lips & \(\{\texttt{mouth}, \texttt{face}\}\) \\
Buckle button, pulling shirt collar, adjusting tie & \(\{\texttt{body-hand}\}\) & Arms akimbo & \(\{\texttt{arms}\}\) \\
Touching or covering suprasternal notch & \(\{\texttt{body-hand}, \texttt{neck}\}\) & Shaking shoulders & \(\{\texttt{shoulders}, \texttt{torso}\}\) \\
Scratching back & \(\{\texttt{body-hand}\}\) & Illustrative BLs & \(\varnothing\) \\
\bottomrule
\end{tabular}
\end{table*}

\subsection{Prompt Design} 
\label{appendix:prompt_design}
\paragraph{\textbf{MGR Prompt.}}
In this work, we did not invest much effort in prompt tuning. We simply use the two base prompts for iMiGUE and MA-52 datasets in all free-form generation experiments. The prompt for iMiGUE is listed below. Here, \texttt{\{labels\}} denotes the full label set of the target dataset. 
    \begin{quote}
    \small
    \ttfamily
    Which micro-gesture best describes the person in this video clip? When deciding, pay attention to visible cues from: hands, fingers, face, mouth, eyes, head, neck, shoulders, arms, and torso.\par
    
    Now do TOP-5 instead of TOP-1:\par
    Return exactly 5 UNIQUE labels from the provided LABEL SET, ranked from most likely (\#1) to less likely (\#5).\par
    Each label must be an EXACT string from LABEL SET. Do not output anything except the JSON object below.\par
    Before outputting, verify that ``labels'' has exactly 5 unique non-empty strings. If not, fix it.\par
    
    LABEL SET:\par
    \{labels\}\par
    
    Output JSON only:\par
    \{\par
    \hspace*{1.5em}"sequence": "video clip",\par
    \hspace*{1.5em}"labels": ["\textless rank1\textgreater", "\textless rank2\textgreater", "\textless rank3\textgreater", "\textless rank4\textgreater", "\textless rank5\textgreater"]\par
    \}\par
    \end{quote}
The prompt for MA-52 is analogous, with minor dataset-specific modifications.
    \begin{quote}
    \small
    \ttfamily
    Which micro-movement best describes the person in this video clip? When deciding, pay attention to visible cues from: hands, fingers, face, mouth, eyes, head, neck, shoulders, arms, torso, legs, feet.\par
    
    Now do TOP-5 instead of TOP-1:\par
    Return exactly 5 UNIQUE labels from the provided LABEL SET, ranked from most likely (\#1) to less likely (\#5).\par
    Each label must be an EXACT string from LABEL SET. Do not output anything except the JSON object below.\par
    Before outputting, verify that ``labels'' has exactly 5 unique non-empty strings. If not, fix it.\par
    
    LABEL SET:\par
    \{labels\}\par
    
    Output JSON only:\par
    \{\par
    \hspace*{1.5em}"sequence": "video clip",\par
    \hspace*{1.5em}"labels": ["\textless rank1\textgreater", "\textless rank2\textgreater", "\textless rank3\textgreater", "\textless rank4\textgreater", "\textless rank5\textgreater"]\par
    \}\par
    \end{quote}

Notably,to encourage the MLLM to return exactly five unique labels from the predefined label set and to reduce misaligned outputs, we explicitly impose these constraints in the prompt.

For NLL-based inference, we use a semantically aligned prompt but remove these output-format requirements. The MGR prompt will be combined with a wrapper prompt before being fed to the MLLM.
    \begin{quote}
    \small
    \ttfamily
    Which micro-gesture best describes the person in this video clip? When deciding, pay attention to visible cues from: hands, fingers, face, mouth, eyes, head, neck, shoulders, arms, and torso.\par
    
    Candidate labels:\par
    \{labels\}\par
    Respond with exact one label from the candidate list. \par
    \end{quote}
However, we observe that if the label list is not explicitly added in the prompt, the results of NLL-based Qwen2.5-VL become substantially different, as shown in Table~\ref{table:appendix_comparison_between_with_or_without_prompt}. If the label list is explicitly added, the prediction-collapse problem will be amplified. In this context, we choose the prompt variant that yields lower performance but remains semantically closer to the free-form generation setting.

\paragraph{\textbf{Tree Search Prompt.}} In tree search, there are three question prompts and two wrapper prompts which are combined with the question prompts, as discussed in SM~\ref{appendix:implementation_details_of_zoom_search}. These prompts differ between Qwen2.5-VL and the other backbones, since the models exhibit different sensitivities to prompt wording. All of them are listed in Table~\ref{table:appendix_other_prompts}.

\section{Complete Analysis of MLLM Failures in MGR}  
\subsection{Analysis of Free-Form Generation Results} 
\label{appendix:complete_analysis_of_MLLMs_failures_in_MGR}
This section provides a failure case of free-form generation, while the correct prediction is obtained by additionally providing the local evidence.

\begin{figure}[t]
    \centering
    \begin{subfigure}[t]{0.95\linewidth}
        \centering
        \includegraphics[width=0.9\linewidth]{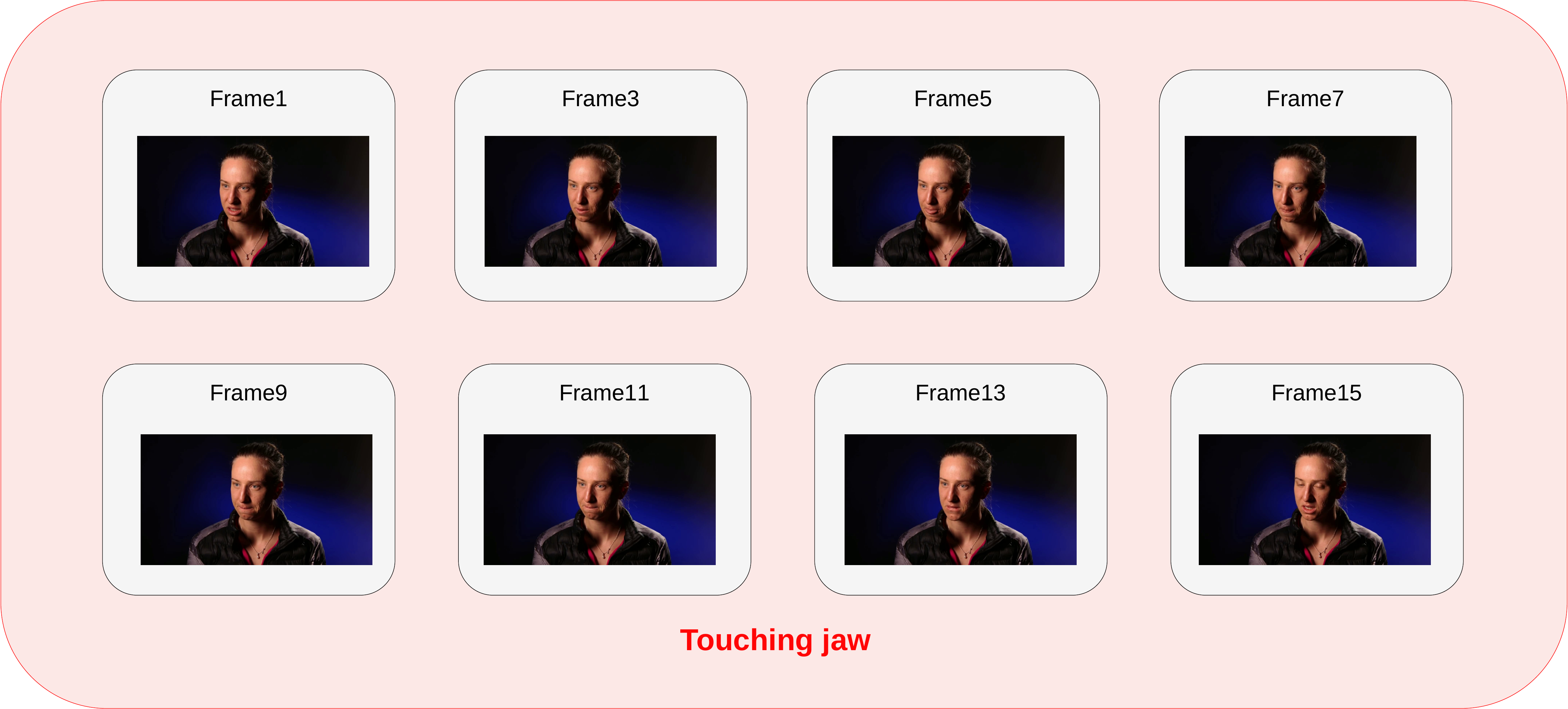}
        \caption{}
        \label{fig:free_form_analysis_case1}
    \end{subfigure}
    \begin{subfigure}[t]{0.95\linewidth}
        \centering
        \includegraphics[width=0.9\linewidth]{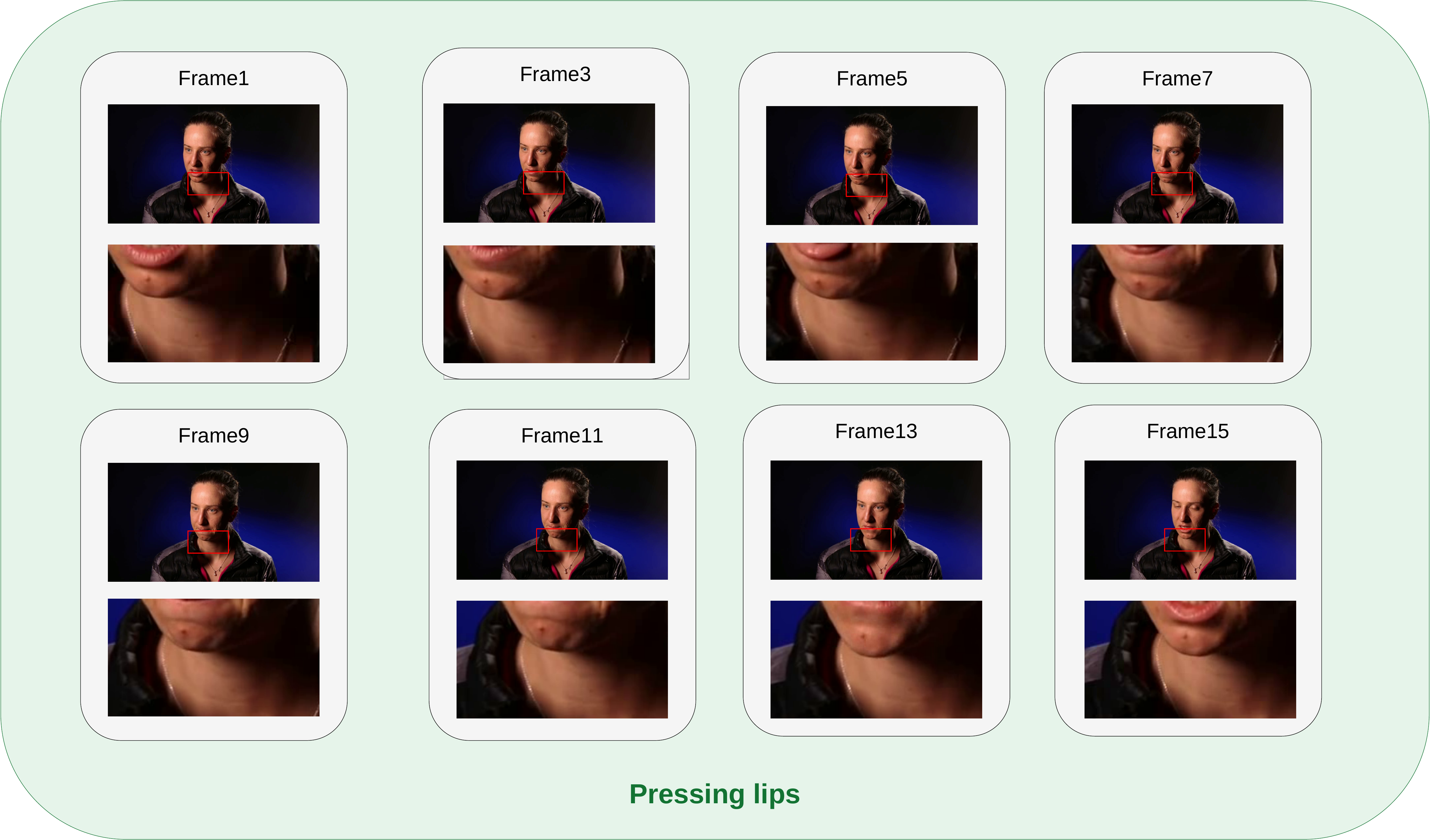}
        \caption{}
        \label{fig:free_form_analysis_case2}
    \end{subfigure}
    \caption{A failure case of free-form generation, analogous to Fig.~3. The GT of this sample is \textit{Pressing lips}. \textbf{(a)}: When only the original video clip is provided, the vanilla MLLM predicts \textit{Touching jaw}. \textbf{(b)}: When additional local evidence of the mouth is provided, the vanilla MLLM will predict \textit{Pressing lips}.}
    \label{fig:free_form_analysis_case_study}
\end{figure}

\subsection{Analysis of NLL-Based Label Scoring Results} 
\label{appendix:analysis_nll_based}

To further quantify the second observation in Sec.~4.2, we analyze the NLL-based label scoring results in the bias-diagnosis space defined in SM~\ref{appendix:detailed_derivation_calibration}. Specifically, $A_{\mathrm{blank}}(x)$ and $A_{\mathrm{lowrank}}(x)$ measure how strongly the blank and lowrank references favor the baseline wrong label $y_w$ over the ground-truth label $y_t$, respectively. Therefore, a positive $A_{\mathrm{blank}}$ indicates language prior bias, while a positive $A_{\mathrm{lowrank}}$ indicates appearance bias beyond the blank reference. Since the calibration subtracts the corresponding reference ($x_b$/$x_r$) score, larger positive values imply that the corresponding calibration method removes stronger wrong-label support.

\begin{figure}[t]
    \centering
    \includegraphics[width=0.85\linewidth]{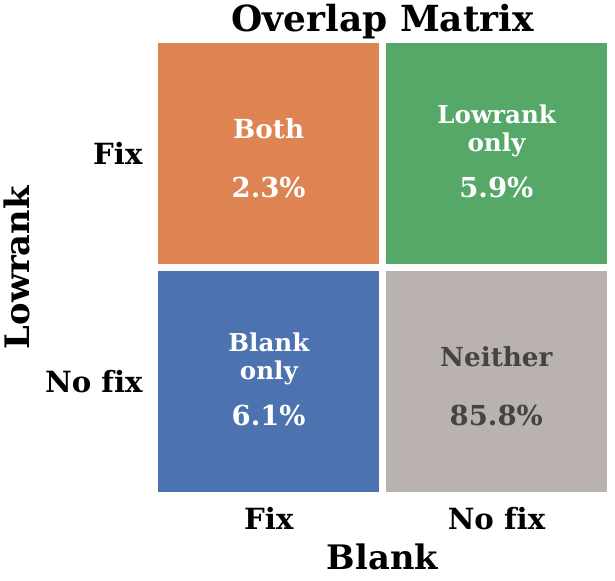}
    \caption{Sample-level overlap of the two calibrations on vanilla-wrong samples. \emph{Both} means that the sample is corrected by both blank and lowrank calibration; \emph{Blank only} and \emph{Lowrank only} mean that only one calibration corrects it; \emph{Neither} means that neither method corrects it. Each cell reports the corresponding percentage within the vanilla-wrong subset.}
    \label{fig:nll_based_analysis_b}
\end{figure}

At the sample level, Fig.~\ref{fig:nll_based_analysis_b} shows that the two calibrations correct overlapping yet clearly distinct subsets of errors. Among the baseline-wrong samples, blank calibration solely corrects 6.1\%, lowrank calibration corrects 5.9\% on its own, and both calibrations correct 2.3\%, while 85.8\% remain wrong. Accordingly, blank and lowrank calibrations correct 8.4\% and 8.2\% of the baseline-wrong samples in total, respectively. However, only 16.1\% of the corrected samples are shared by the two branches. This quantitatively supports the statement in Sec.~4.2 that the two calibrations are complementary rather than redundant. At the same time, the large unresolved portion shows that bias correction alone is insufficient to solve zero-shot MGR, which is consistent with the need for localized evidence acquisition.

\begin{figure*}[t]
    \centering
    \includegraphics[width=\linewidth]{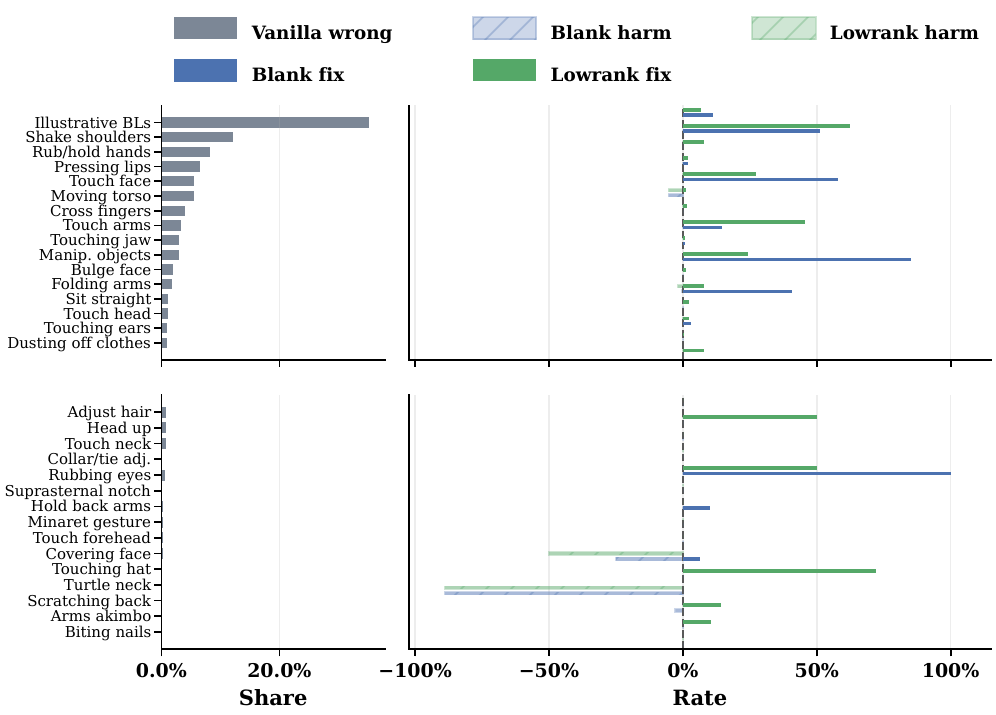}
    \caption{Class-wise effect of blank and lowrank calibration in NLL-based label scoring. Left: the ground-truth class share of samples that are misclassified by the vanilla model (\emph{vanilla wrong}). Right: per-class calibration effect, where \emph{fix} means that a sample is originally misclassified by the vanilla model but becomes correct after the corresponding calibration method, and \emph{harm} means that a sample is originally correct under the vanilla model but becomes wrong after calibration. The rates are normalized by class size.}
    \label{fig:nll_based_analysis_a}
\end{figure*}

\begin{figure}[t]
    \centering
    \includegraphics[width=\linewidth]{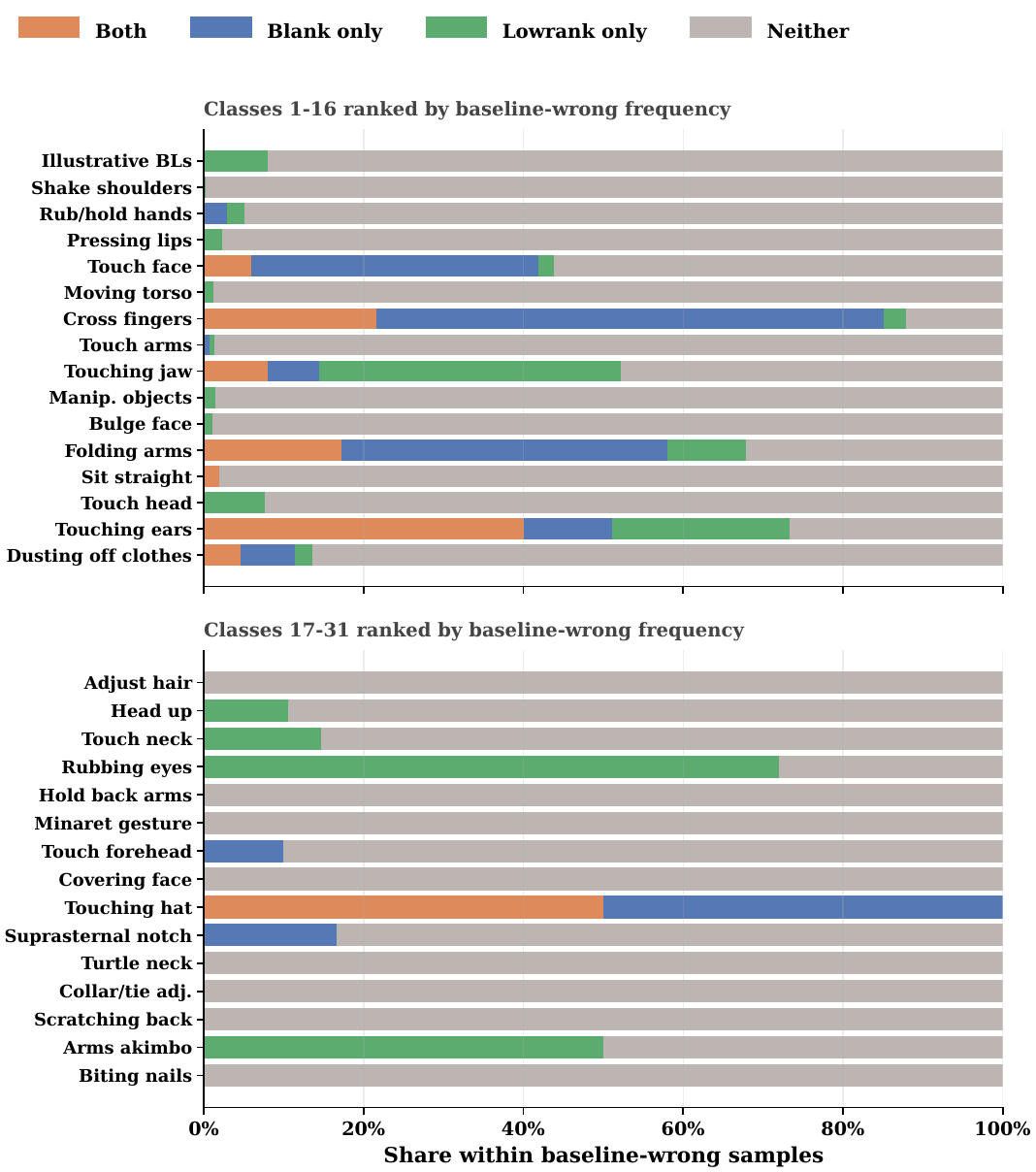}
    \caption{Per-class overlap composition on vanilla-wrong samples. For each class, the bar is decomposed into four groups: \emph{Both}, \emph{Blank only}, \emph{Lowrank only}, and \emph{Neither}, using the same definitions as in Fig.~\ref{fig:nll_based_analysis_b}. Thus, each bar shows how the two calibrations overlap, or fail to overlap, when correcting errors in that class.}
    \label{fig:nll_based_analysis_c_bar}
\end{figure}

Figs.~\ref{fig:nll_based_analysis_a} and \ref{fig:nll_based_analysis_c_bar} further show that this complementarity is strongly class-dependent under the long-tailed label distribution. Blank calibration is particularly effective on classes whose errors are more strongly driven by the language prior, such as \textit{Crossing fingers} (85.1\% of baseline-wrong samples corrected by blank calibration), \textit{Folding arms} (58.0\%), and \textit{Touching or scratching facial parts} (41.9\%). In contrast, lowrank calibration is relatively stronger on classes where static appearance induces a different bias pattern, such as \textit{Rubbing eyes} (72.0\%), \textit{Touching jaw} (45.7\%), \textit{Touching ears} (62.2\%), and even \textit{Illustrative BLs} (8.0\%) in the most error-dominant class. For each class, we compare blank and lowrank calibration by their class-wise correction effect, defined as the corrected fraction minus the harmed fraction within that class. By this criterion, lowrank calibration is better on 41.9\% of the classes, blank calibration is better on 25.8\%, and they are comparable on the remaining 32.3\%. Fig.~\ref{fig:nll_based_analysis_a} also shows that over-correction is concentrated in a few rare categories with very small baseline-correct subsets, rather than being the dominant pattern. Therefore, the main signal is not a uniform advantage of one calibration over the other, but rather a class-dependent trade-off between the two.

\begin{figure}[t]
    \centering
    \begin{subfigure}[t]{0.95\linewidth}
        \centering
        \includegraphics[width=0.9\linewidth]{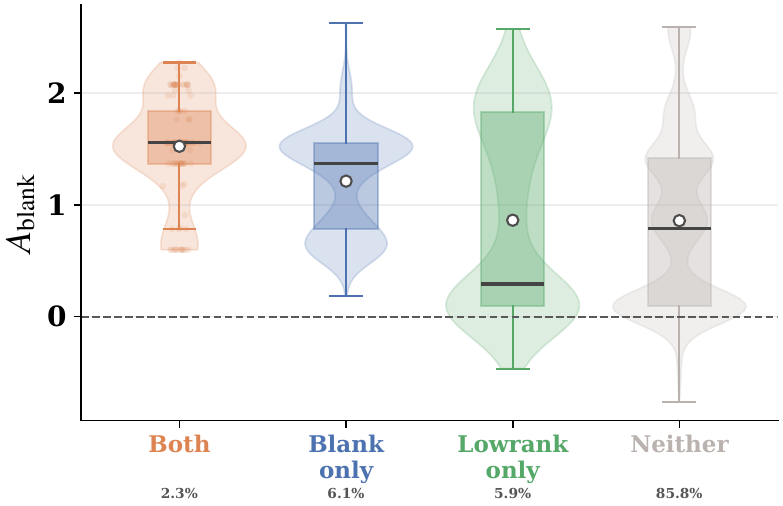}
        \caption{Distribution of $A_{\mathrm{blank}}$ across the four overlap groups.}
        \label{fig:nll_based_analysis_d_blank}
    \end{subfigure}
    \begin{subfigure}[t]{0.95\linewidth}
        \centering
        \includegraphics[width=0.9\linewidth]{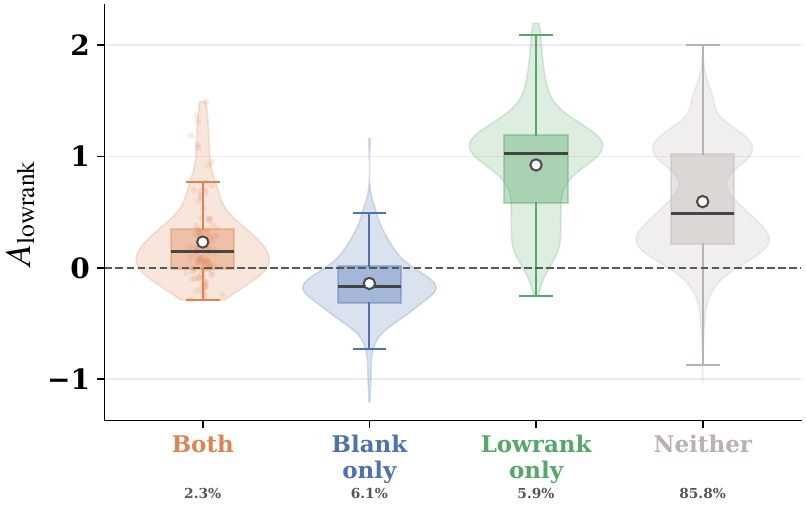}
        \caption{Distribution of $A_{\mathrm{lowrank}}$ across the four overlap groups.}
        \label{fig:nll_based_analysis_d_lowrank}
    \end{subfigure}
    \caption{Bias diagnosis by overlap group. The four groups (\emph{Both}, \emph{Blank only}, \emph{Lowrank only}, and \emph{Neither}) are defined in the same way as in Fig.~\ref{fig:nll_based_analysis_b}. $A_{\mathrm{blank}}$ measures language-prior bias revealed by the blank reference, while $A_{\mathrm{lowrank}}$ measures appearance bias revealed by the lowrank reference relative to the blank reference.}
    \label{fig:nll_based_analysis_d}
\end{figure}

The mechanism plots in Fig.~\ref{fig:nll_based_analysis_d_blank} and Fig.~\ref{fig:nll_based_analysis_d_lowrank} help explain why the two calibrations behave differently. For samples corrected only by blank calibration, $A_{\mathrm{blank}}$ is clearly positive and large (mean 1.21, median 1.37), whereas $A_{\mathrm{lowrank}}$ is slightly negative on average (mean -0.14, median -0.17). This indicates that these failures are mainly dominated by language prior bias, while the appearance-bias term does not provide additional wrong-label preference. By contrast, lowrank-only fixes have a much larger positive $A_{\mathrm{lowrank}}$ (mean 0.92, median 1.03) together with a smaller and more dispersed $A_{\mathrm{blank}}$ (mean 0.86, median 0.29), showing that these errors are better explained by appearance bias exposed by the lowrank reference. Samples corrected by both calibrations have the largest $A_{\mathrm{blank}}$ (mean 1.53) and still positive $A_{\mathrm{lowrank}}$ (mean 0.23), indicating that some failures simultaneously contain both language and appearance bias. Even the unresolved group remains positive on average in both quantities ($A_{\mathrm{blank}}=0.86$, $A_{\mathrm{lowrank}}=0.59$), suggesting that residual bias is still present, although bias correction alone is often insufficient to reverse the ranking.

Overall, the quantitative diagnosis in this section is consistent with the second paragraph of Sec.~4.2. Vanilla MLLMs suffer not only from insufficient localized evidence, but also from two partially separable prior components in label scoring: language-prior bias revealed by the blank calibration and appearance bias revealed by the lowrank calibration. Because these two components are only partially overlapping and strongly class-dependent, the calibration-aggregating formulation in Sec.~3.4 is better justified than choosing either reference alone. This also explains why lowrank-only calibration can be weaker than blank-only calibration in overall performance, yet still remains useful after aggregation, since it corrects a different subset of samples and classes.

\begin{table}[t]
    \centering
    \caption{Comparison of NLL-based Qwen2.5-VL with and without the label list in the prompt.}
    \setlength{\tabcolsep}{3pt}
    \begin{tabular}{@{}l ccc@{}}
        \toprule
        \textbf{Method} & \textbf{mCA@1} & \textbf{mCA@5} & \textbf{MF1} \\
        \midrule
        Qwen2.5-VL & 0.1667 & 0.3561 & 0.0563 \\
        Qwen2.5-VL w/ label list & 0.0523 & 0.2524 & 0.0072 \\
        \bottomrule
    \end{tabular}
    \label{table:appendix_comparison_between_with_or_without_prompt}
\end{table}

\section{NLL-Based Label Scoring vs. Prompt-Based Free-Form Generation} 
\label{appendix:nll_based_vs_prompt_based}
This section complements Sec.~4.3 in two aspects. First, we revisit why the raw NLL-based scoring results in Table~1 in the main paper are often weaker than prompt-based free-form generation under the current aligned prompt design. Our goal here is not to explain the internal mechanism of blank calibration. Instead, we use blank calibration as an intervention: if subtracting the blank reference substantially reduces prediction concentration and partially restores performance, then the weak NLL-based scoring results are more plausibly attributed to prior-driven score collapse than to a complete absence of useful visual evidence. Second, we revisit the dataset-related reversal observed in Table~1 in the main paper under raw inference.

For a label set \(Y\), let \(p_{\mathrm{Top1}}(y)\) denote the Top-1 prediction histogram over \(Y\). To quantify prediction concentration beyond mCA@1, we additionally report the number of distinct Top-1 labels, the dominant-label ratio \(\max\limits_y p_{\mathrm{Top1}}(y)\), the normalized entropy
\begin{equation}
H_{\mathrm{norm}}
=
-\frac{1}{\log |Y|}
\sum_{y\in Y}
p_{\mathrm{Top1}}(y)\log p_{\mathrm{Top1}}(y),
\end{equation}
and the Jensen--Shannon divergence (JSD) between the Top-1 prediction histogram and the ground-truth label histogram. Here, lower \(H_{\mathrm{norm}}\) indicates a more collapsed Top-1 prediction distribution with poorer label-space coverage, whereas larger JSD indicates a larger global discrepancy between the model's aggregate Top-1 preference and the GT label histogram.

\begin{table*}[t]
\centering
\caption{iMiGUE prediction-distribution comparison among prompt-based free-form generation, raw NLL-based scoring, and blank-calibrated NLL-based scoring. For prompt-based and raw NLL-based, mCA@1 is aligned with Table~1 in the main paper; the remaining statistics are computed from the corresponding Top-1 histograms. Lower dominant ratio and JSD are better, while higher \#Top-1 labels and \(H_{\mathrm{norm}}\) indicate less concentrated predictions.}
\label{tab:sm_imigue_blankcal}
\resizebox{\textwidth}{!}{
\begin{tabular}{lllccccc l}
\toprule
Dataset & Model & Mode & mCA@1 & \#Top-1 labels & Dominant ratio & $H_{\mathrm{norm}}$ & JSD & Dominant Top-1 label \\
\midrule
iMiGUE & Qwen2.5-VL-7B & Prompt-based & 0.1615 & 19 & 37.6\% & 0.503 & 0.478 & Touching or scratching forehead \\
iMiGUE & Qwen2.5-VL-7B & Vanilla NLL & 0.0523 & 6 & 89.9\% & 0.116 & 0.643 & Buckle button, pulling shirt collar, adjusting tie \\
iMiGUE & Qwen2.5-VL-7B & Blank-cal. NLL & 0.1363 & 23 & 38.6\% & 0.614 & 0.430 & Crossing fingers \\
\midrule
iMiGUE & MiMo-VL-7B & Prompt-based & 0.2445 & 27 & 54.1\% & 0.491 & 0.339 & Rubbing or holding hands \\
iMiGUE & MiMo-VL-7B & Vanilla NLL & 0.0325 & 2 & 99.7\% & 0.006 & 0.673 & Buckle button, pulling shirt collar, adjusting tie \\
iMiGUE & MiMo-VL-7B & Blank-cal. NLL & 0.1573 & 26 & 28.9\% & 0.623 & 0.310 & Touching or scratching facial parts \\
\midrule
iMiGUE & InternVL3.5-8B & Prompt-based & 0.2034 & 22 & 48.2\% & 0.531 & 0.388 & Rubbing or holding hands \\
iMiGUE & InternVL3.5-8B & Vanilla NLL & 0.0329 & 2 & 75.5\% & 0.161 & 0.666 & Buckle button, pulling shirt collar, adjusting tie \\
iMiGUE & InternVL3.5-8B & Blank-cal. NLL & 0.1768 & 21 & 25.1\% & 0.598 & 0.337 & Touching jaw \\
\bottomrule
\end{tabular}
}
\end{table*}

\paragraph{\textbf{A More Severe Prediction-Collapse Problem in Raw NLL-Based Label Scoring.}}
The main paper already shows that prompt-based free-form generation itself is not free from prediction collapse on iMiGUE. However, Table~\ref{tab:sm_imigue_blankcal} shows that under the aligned label-list prompt, the raw NLL-based scoring amplifies this effect much more severely. Compared with prompt-based generation, raw NLL-based scoring produces far fewer distinct Top-1 labels, much larger dominant-label ratios, lower normalized entropy, and larger JSD values for all three backbones. Moreover, the dominant Top-1 label under raw NLL-based scoring collapses to the same sink label, \emph{Buckle button, pulling shirt collar, adjusting tie}, for all three backbones. Therefore, the weaker NLL-based scoring performance in Table~1 in the main paper is better understood as a consequence of more severe prediction collapse, rather than simply as a generic inferiority of the NLL scoring path.

\paragraph{\textbf{Blank Calibration Alleviates the Collapse.}}
Blank calibration does not serve as the object of explanation in this section; instead, it acts as an intervention that tests whether the raw NLL weakness is prior-driven. As shown in Table~\ref{tab:sm_imigue_blankcal}, after blank calibration, the Top-1 distributions become substantially less concentrated for all three backbones: the number of distinct Top-1 labels increases, the dominant-label ratio drops sharply, the normalized entropy rises, and the JSD to the ground-truth distribution consistently decreases. Meanwhile, mCA@1 also recovers substantially. Since the blank reference suppresses visual content and mainly reflects language prior, this consistent reduction in prediction concentration supports the interpretation that a major part of the weak NLL-based scoring performance comes from prior-driven score collapse.

\paragraph{\textbf{Dataset-Related Reversal under Raw Inference.}}
The dataset-related reversal mentioned in Sec.~4.3 is better discussed under raw inference only. As shown in Table~1 in the main paper, prompt-based free-form generation is consistently stronger on iMiGUE than on MA-52 for all three compared backbones, whereas NLL-based scoring shows the opposite cross-dataset preference, i.e., it is consistently stronger on MA-52 than on iMiGUE. This counterintuitive reversal should not be interpreted as a simple ranking of visual difficulty between the two datasets. A more plausible explanation is that raw NLL ranking is strongly modulated by dataset-dependent prior structure in the label space. Concretely, on iMiGUE the three backbones collapse toward the same clothing-related sink label, whereas on MA-52 the dominant labels shift to more posture- or body-configuration-like categories such as \emph{Sitting straightly} and \emph{Arms akimbo}. This interpretation is also consistent with the ground-truth label histograms, where iMiGUE is substantially more imbalanced than MA-52. Therefore, the reversal is more appropriately understood as a change in collapse pattern and prior structure across datasets.

\paragraph{Prompt-sensitivity note.}
For fairness, the NLL-based prompt is chosen to be semantically aligned with the prompt-based setting. However, the main paper already shows that NLL-based scoring is sensitive to this prompt choice when the candidate label list is included. Therefore, the weaker raw NLL results reported here should be interpreted as an observation under the current aligned prompt design, rather than as evidence that NLL-based scoring is intrinsically inferior in all settings.

\section{Full Results of the Comparison between Zero-MELO and Supervised Discriminative Models} 
\label{appendix:zeroshot_vs_supervised}
We further report the results of supervised discriminative models for reference. As shown in Table~2 in the main paper, Zero-MELO surpasses all variants of BlockerGCN and remains competitive in terms of mCA@1. The gaps to MMN(J) and MMN(2s) are relatively small, whereas PoseC3D remains clearly stronger. Although the MF1 of Zero-MELO remains weaker than that of supervised discriminative models, relatively low MF1 is also commonly observed among MLLM-based methods in Table~1 in the main paper.

\begin{table}[t]
    \centering
    \caption{Full comparison between Zero-MELO and supervised discriminative models on the iMiGUE dataset.}
    \setlength{\tabcolsep}{3pt}
    \begin{tabular}{@{}l ccc@{}}
        \toprule
        \textbf{Method} & \textbf{mCA@1} & \textbf{mCA@5} & \textbf{MF1} \\
        \midrule
        BlockerGCN(J) & 0.2324 & 0.6080 & 0.2375  \\ 
        BlockerGCN(B) & 0.2362 & 0.6040 & 0.2407  \\ 
        BlockerGCN(V) & 0.2375 & 0.5820 & 0.2467  \\ 
        BlockerGCN(3s) & 0.2658 & 0.6469 & 0.2809  \\ 
        PoseC3D & \textbf{0.3402} & \textbf{0.7159} & \textbf{0.3583}  \\
        MMN(J) & 0.2991 & 0.6626 & 0.3062\\
        MMN(B) & 0.2335 & 0.5953 & 0.2428\\
        MMN(2s) & 0.2890 & 0.6565 & 0.3008\\
        \midrule
        Zero-MELO (Ours) & 0.2684 & 0.5775 & 0.1325 \\
        \bottomrule
    \end{tabular}
    \label{table:full_comparsion_with_supervised}
\end{table}

\section{Experiments across Different Backbones} 
\label{appendix:different_backbones}
We further evaluate Zero-MELO across different backbones. Sepcifically, InternVL3.5 and MiMo-VL are implemented on iMiGUE. As shown in Tables~\ref{table:appendix_internvl35_imigue} and~\ref{table:appendix_mimiovl_imigue}, Zero-MELO consistently achieves better results on both backbones. In particular, mCA@1 improves from 20.34\% to 24.25\% on InternVL3.5 and from 24.45\% to 29.77\% on MiMo-VL, while mCA@5 and MF1 are also slightly improved. This suggests that the advantage of Zero-MELO is not limited to a single MLLM backbone.

\begin{table}[t]
    \centering
    \caption{Comparison between Zero-MELO and the corresponding vanilla MLLM under the InternVL3.5 backbone on iMiGUE.}
    \begin{tabular}{@{}l ccc@{}}
        \toprule
        \multirow{2}{*}{\textbf{InternVL3.5}} & \multicolumn{3}{c}{\textbf{iMiGUE}} \\
        &\textbf{mCA@1} & \textbf{mCA@5} & \textbf{MF1} \\
        \midrule
        Prompt-based zero-shot   & 0.2034 & 0.4358 & 0.0927 \\
        NLL-based & 0.0329 & 0.2408 & 0.0007 \\
        Zero-MELO (Ours) & \textbf{0.2425} & \textbf{0.4375} & \textbf{0.0974}  \\
        \bottomrule
    \end{tabular}
    \label{table:appendix_internvl35_imigue}
\end{table}
\begin{table}[t]
    \centering
    \caption{Comparison between Zero-MELO and the corresponding vanilla MLLM under the MiMo-VL backbone on iMiGUE.}
    \begin{tabular}{@{}l ccc@{}}
        \toprule
        \multirow{2}{*}{\textbf{MiMo-VL}} & \multicolumn{3}{c}{\textbf{iMiGUE}} \\
        &\textbf{mCA@1} & \textbf{mCA@5} & \textbf{MF1} \\
        \midrule
        Prompt-based zero-shot   & 0.2445 & 0.5643 & 0.1506 \\
        NLL-based & 0.0325 & 0.2480 & 0.0008 \\
        Zero-MELO (Ours) &\textbf{0.2977} &\textbf{0.5834} & \textbf{0.1534}  \\
        \bottomrule
    \end{tabular}
    \label{table:appendix_mimiovl_imigue}
\end{table}
\section{Experiments across Different Calibration Strategies} 
\label{appendix:different_calibrations}
We ablate the proposed calibration strategies, namely blank calibration, lowrank calibration, and their aggregation. Notably, the verbalizer marginalization is introduced in these experiments, and thus the results in this section are not the same as them in SM~\ref{appendix:nll_based_vs_prompt_based}. As shown in Table~\ref{table:calibration_strategies}, the vanilla model performs poorly, indicating that Qwen2.5-VL is overly affected by dominant biases. Using only the blank calibration already yields a large improvement (mCA@1=0.2040, mCA@5=0.4943, MF1=0.0855), while the lowrank calibration is weaker (mCA@1=0.1455, mCA@5=0.3717, MF1=0.0569). This is also consistent with SM~\ref{appendix:analysis_nll_based}. The best results are obtained by aggregating the blank and lowrank calibrations through the aggregation (mCA@1=0.2363, mCA@5=0.5436, MF1=0.1097). This indicates that the two calibrations provide complementary information: the blank calibration offers stronger robustness to language-prior bias, whereas the lowrank calibration contributes additional discriminative structure. Their aggregation produces the most reliable label scores.
\begin{table}[t]
    \centering
    \caption{Comparison experiments across different calibration strategies.}
    \begin{tabular}{@{}l ccc@{}}
        \toprule
        \textbf{Method} & \textbf{mCA@1} & \textbf{mCA@5} & \textbf{MF1} \\
        \midrule
        Vanilla  & 0.0357 & 0.2084 & 0.0094\\
        + Cali (Lowrank)  & 0.1455 & 0.3717 & 0.0569\\
        + Cali (Blank) & 0.2040 & 0.4943 & 0.0855\\
        + Cali (Aggregation) & \textbf{0.2363} & \textbf{0.5436} & \textbf{0.1097}\\ 
        \bottomrule
    \end{tabular}
    \label{table:calibration_strategies}
\end{table}

\section{Detailed Implementations of Component-Validity Ablations} 
\label{appendix:component_validity}

In Sec.~4.4, the contributions of the major components of Zero-MELO are studied under a unified evaluation setup. All results use the same label set, the same tree-searched cue-guided views, the same label-conditioned cue compatibility setting introduced in Sec.~3.5, and the same verbalizer marginalization coefficient. The difference lies only in how the final label score is constructed.

\paragraph{\textbf{Vanilla.}}
This implementation uses the raw global NLL \(\ell_g(y)\) directly, without calibration or local evidence. The difference between this implementation and the pure NLL-based vanilla baseline lies in the constrained input format. The pure NLL-based baseline uses a single global video input,
\begin{equation}
    y^{*}_{NLL}=\arg\min_{y\in\mathcal{Y}}\ell(y\mid g, t_{NLL}),
\end{equation}
where $t_{NLL}$ indicates the normal prompt stated in SM~\ref{appendix:prompt_design}. By contrast, this vanilla implementation reformulates the original global clip into the same paired-input format used by each cue branch. Specifically, the prediction is given by
\begin{equation}
    y^{*}_{vanilla}=\arg\min_{y\in\mathcal{Y}}\ell(y\mid (g,b), t_{zero-melo})
\end{equation}
where $b$ is a blank view used as a placeholder, and $t_{zero-melo}$ denotes the combined prompt formed by the zoom wrapper prompt and $t_{NLL}$. This constrained global input is used to align the input format with that of each cue branch, so that no extra bias is introduced.

\paragraph{\textbf{+ TS.}}
This implementation introduces tree-searched local evidence without applying test-time calibration. It uses the global branch as an anchor and refines it with the most supportive compatible local cue branch. 

\paragraph{\textbf{+ Cali (Blank).}}
This implementation uses only the global score with blank calibration, in order to test whether suppressing the language prior alone is sufficient. Result of this calibration is not aligned with that in Table~\ref{tab:sm_imigue_blankcal}, since the verbalizer marginalization technique is leveraged.

\paragraph{\textbf{+ Cali (Aggregate).}}
This implementation uses the dual-reference calibrated global score in Sec.~3.4, which aggregates the results of blank and lowrank calibrations.

\paragraph{\textbf{+ TS + Cali (Blank).}}
This implementation combines tree-searched local evidence with blank-only calibrated global scores, but does not yet use the final multi-cue fusion rule. It is worth noting that the \textbf{+TS} implementation is not expected to perform strongly on its own since the visual modality is overwhelmed by the language-prior bias. Notably, tree search only produces a set of candidate local views, but in the zero-shot setting, there is no GT prior indicating which single cue branch should be selected as the decisive local evidence for the unknown target label. As a result, single-cue inference after tree search is inherently underdetermined. This is precisely the reason why Zero-MELO introduces a multi-cue fusion module: instead of committing to one potentially unreliable local branch, it fuses multiple compatible local evidence sources on top of the calibrated global anchor.

\paragraph{\textbf{+ TS + Cali + F(Max).}}
This implementation corresponds to a cue-expert variant, in which the final decision is determined by the most supportive calibrated cue branch over these compatible cue sets. The difference between \textbf{+ TS + Cali} and this ablation is that the former is still a single-cue refinement baseline, while the latter is already a multi-cue fusion baseline. Specifically, \textbf{+ TS + Cali (Blank)} uses the blank-calibrated global branch as an anchor and refines it with only the most supportive compatible blank-calibrated cue branch. In contrast, \textbf{+ TS + Cali + F(Max)} jointly considers the aggregated calibrated global branch and all compatible calibrated cue branches, and performs label-wise expert selection by taking the lowest calibrated score among them.

\paragraph{\textbf{+ TS + Cali + F(Average).}}
This implementation corresponds to a uniform fusion variant, in which the calibrated global branch and all compatible calibrated cue branches are averaged with equal weight.

\paragraph{\textbf{+ TS + Cali + F(Ours).}}
This implementation is the full method in Sec.~3.5, namely Label-Conditioned Normalized Fusion, which combines the calibrated global branch and the compatible calibrated cue branches using label-conditioned cue selection and family-dependent weighting.

In this way, the ablation progressively reveals the effects of local evidence acquisition, test-time calibration, and the final multi-cue fusion rule.

\section{Open-Vocabulary Extension and Transferability}
To verify generalization, we further transfer Zero-MELO,
without any manual design, directly to a general action recognition
benchmark, NTU120-110/10, using automatically generated associations. As shown in Table~\ref{table:NTU120-10}, Zero-MELO outperforms vanilla Qwen2.5-VL-7B (0.663→0.694) without any tuning, while remaining competitive/even stronger than other listed methods use additional training data and extra modalities (e.g., R+T+S).
\begin{table}[t]
    \centering
    \caption{Directly transfer Zero-MELO to the zeroshot task on NTU120-110/10 without manual intervention. \textbf{M}: modality. \textbf{Train}: trained on seen split. ACC@1 is reported for consistency with other methods. R/T/S denote RGB/text/skeleton.}
    \label{table:NTU120-10}
    
    \small 
    \setlength{\tabcolsep}{2pt} 
    
    \resizebox{\linewidth}{!}{%
        \begin{tabular}{@{}lccc | lccc@{}}
            \toprule
            \textbf{Method} & \textbf{Acc@1} & \textbf{M} & \textbf{Train} & \textbf{Method} & \textbf{Acc@1} & \textbf{M} & \textbf{Train} \\
            \midrule 
            SynSE [ICIP'21]  & 0.576 & S & \checkmark & SKI-FROSTER [AAAI'25]  & 0.685 & R+T+S & \checkmark \\
            FS-VAE [ICCV'25] & 0.679 & S & \checkmark & SKI-ViFiCLIP [AAAI'25] & \textbf{0.775} & R+T+S & \checkmark \\
            SkeletonContext [CVPR'26]  & 0.655  & S+T & \checkmark & Qwen2.5-VL-7B [5] & 0.663 & R & $\times$ \\
            XCLIP [ECCV'22] & 0.578 & R+T & $\times$ & Zero-MELO (\textbf{Ours}) & \underline{0.694} & R & $\times$ \\
            \bottomrule
        \end{tabular}%
    } 
\end{table}

\section{Runtime Analysis}
Runtime is measured on the A100 40GB, as shown in Table~\ref{table:runtime_result}. We implement KV cache and batch inference for acceleration, as mentioned in Sec.~4.6. We acknowledge there is a trade-off between the accuracy and runtime, which we will improve in future work.
\begin{table}[t]
    \centering
    \caption{Average runtime per iMiGUE sample. \textbf{Time} is in seconds, and \textbf{Passes} denotes MLLM calls.}
    \label{table:runtime_result}
    \vspace{-0.35cm} 
    
    \small 
    \setlength{\tabcolsep}{3pt} 
    
    \resizebox{\linewidth}{!}{%
        \begin{tabular}{@{}lcc | lcc@{}}
            \toprule
            \textbf{Method} & \textbf{Time} & \textbf{Passes} & \textbf{Method} & \textbf{Time} & \textbf{Passes} \\ 
            \midrule
            Prompt-based Qwen2.5-VL & 2.71 & 1   & 
            Zero-MELO (\textbf{Ours}) & 399.82 & 585 \\
            NLL-based Qwen2.5-VL & 17.12 & 96    & 
            - TS & 389.63 & 510 \\
            \cmidrule{1-3} 
            ZoomEye [47] & 325.29 & 690          & 
            - Cali & 9.52 & 75 \\
            - Context Learning & 2.03 & 10       & 
            - Fusion & 0.59 & 0 \\
            - ZoomEye Search & 308.98 & 648      & 
             & & \\
            - Answer & 14.18 & 32                & 
             & & \\
            \bottomrule 
        \end{tabular}%
    } 
\end{table}

\begin{table*}[t]
\centering
\caption{Label-conditioned cue sets \(\mathcal{C}(y)\) for MA-52.}
\label{table:appendix_label_conditioned_cue_set_ma52}
\scriptsize
\setlength{\tabcolsep}{3.5pt}
\renewcommand{\arraystretch}{0.97}
\begin{tabular}{p{0.18\textwidth} p{0.28\textwidth} p{0.18\textwidth} p{0.28\textwidth}}
\toprule
\textbf{Label \(y\)} & \textbf{Label-Conditioned Cue Set \(\mathcal{C}(y)\)} & \textbf{Label \(y\)} & \textbf{Label-Conditioned Cue Set \(\mathcal{C}(y)\)} \\
\midrule
Shaking body & \(\{\texttt{torso}, \texttt{legs}\}\) & Spread legs & \(\{\texttt{legs}\}\) \\
Sitting straightly & \(\{\texttt{torso}\}\) & Closing legs & \(\{\texttt{legs}\}\) \\
Shrugging & \(\{\texttt{torso}\}\) & Crossing legs & \(\{\texttt{legs}\}\) \\
Turning around & \(\{\texttt{torso}, \texttt{legs}\}\) & Stretching feet & \(\{\texttt{feet}, \texttt{legs}\}\) \\
Rising up & \(\{\texttt{torso}, \texttt{legs}\}\) & Retracting feet & \(\{\texttt{feet}, \texttt{legs}\}\) \\
Bowing head & \(\{\texttt{head}, \texttt{neck}\}\) & Tiptoe & \(\{\texttt{feet}, \texttt{legs}\}\) \\
Head up & \(\{\texttt{head}, \texttt{neck}\}\) & Scratching or touching neck & \(\{\texttt{hands}, \texttt{neck}\}\) \\
Tilting head & \(\{\texttt{head}, \texttt{neck}\}\) & Scratching or touching chest & \(\{\texttt{hands}, \texttt{torso}\}\) \\
Turning head & \(\{\texttt{head}, \texttt{neck}\}\) & Scratching or touching back & \(\{\texttt{hands}, \texttt{torso}\}\) \\
Nodding & \(\{\texttt{head}, \texttt{neck}\}\) & Scratching or touching shoulder & \(\{\texttt{hands}\}\) \\
Shaking head & \(\{\texttt{head}, \texttt{neck}\}\) & Arms akimbo & \(\{\texttt{torso}\}\) \\
Scratching arms & \(\{\texttt{hands}, \texttt{hands and arms}\}\) & Crossing arms & \(\{\texttt{torso}\}\) \\
Playing objects & \(\{\texttt{hands}, \texttt{hands and arms}\}\) & Playing or tidying hair & \(\{\texttt{hands}, \texttt{head}, \texttt{head-hand}\}\) \\
Putting hands together & \(\{\texttt{hands}, \texttt{hands and arms}\}\) & Scratching or touching hindbrain & \(\{\texttt{hands}, \texttt{head}, \texttt{head-hand}\}\) \\
Rubbing hands & \(\{\texttt{hands}, \texttt{hands and arms}\}\) & Scratching or touching forehead & \(\{\texttt{hands}, \texttt{head-hand}\}\) \\
Pointing oneself & \(\{\texttt{hands}, \texttt{torso}, \texttt{hands and arms}\}\) & Scratching or touching face & \(\{\texttt{hands}, \texttt{head-hand}\}\) \\
Clenching fist & \(\{\texttt{hands}, \texttt{hands and arms}\}\) & Rubbing eyes & \(\{\texttt{hands}, \texttt{head-hand}\}\) \\
Stretching arms & \(\{\texttt{hands}, \texttt{hands and arms}\}\) & Touching nose & \(\{\texttt{hands}, \texttt{head-hand}\}\) \\
Retracting arms & \(\{\texttt{hands}, \texttt{hands and arms}\}\) & Touching ears & \(\{\texttt{hands}, \texttt{head}, \texttt{head-hand}\}\) \\
Waving & \(\{\texttt{hands}, \texttt{hands and arms}\}\) & Covering face & \(\{\texttt{hands}, \texttt{head-hand}\}\) \\
Spreading hands & \(\{\texttt{hands}, \texttt{hands and arms}\}\) & Covering mouth & \(\{\texttt{hands}, \texttt{mouth}, \texttt{head-hand}\}\) \\
Hands touching fingers & \(\{\texttt{hands}, \texttt{hands and arms}\}\) & Pushing glasses & \(\{\texttt{hands}, \texttt{head-hand}\}\) \\
Other finger movements & \(\{\texttt{hands}, \texttt{hands and arms}\}\) & Patting legs & \(\{\texttt{hands}, \texttt{legs}, \texttt{leg-hand}\}\) \\
Illustrative gestures & \(\{\texttt{hands}, \texttt{hands and arms}\}\) & Touching legs & \(\{\texttt{hands}, \texttt{legs}, \texttt{leg-hand}\}\) \\
Shaking legs & \(\{\texttt{legs}, \texttt{feet}\}\) & Scratching legs & \(\{\texttt{hands}, \texttt{legs}, \texttt{leg-hand}\}\) \\
Curling legs & \(\{\texttt{legs}, \texttt{feet}\}\) & Scratching feet & \(\{\texttt{hands}, \texttt{feet}, \texttt{leg-hand}\}\) \\
\bottomrule
\end{tabular}
\end{table*}

\begin{table*}[t]
\centering
\caption{Prompt templates for \textbf{region cues} used by different backbones. Here, \texttt{\{cue\}} denotes the queried region cue.}
\label{table:appendix_other_prompts}
\small
\setlength{\tabcolsep}{4pt}
\renewcommand{\arraystretch}{1.15}
\begin{tabular}{@{}p{0.16\textwidth}p{0.39\textwidth}p{0.39\textwidth}@{}}
\toprule
\textbf{Prompt type} & \textbf{Qwen2.5-VL} & \textbf{InternVL3.5 / MiMo-VL} \\
\midrule
\multicolumn{3}{@{}c@{}}{\textit{Wrapper Prompts}} \\
\midrule
Global wrapper &
\parbox[t]{\linewidth}{\ttfamily
\textless video\textgreater
}
&
\parbox[t]{\linewidth}{\ttfamily
Main video frames:\newline
\textless video\textgreater
}
\\
Zoom wrapper &
\parbox[t]{\linewidth}{\ttfamily
\textless video\textgreater\newline
This is the main video, and the section enclosed by the red rectangle in each frame is the focus region.\newline
\textless video\textgreater\newline
This is the zoomed-in view of the focus region.
}
&
\parbox[t]{\linewidth}{\ttfamily
Main video frames (the section enclosed by the red rectangle in each frame is the focus region):\newline
\textless video\textgreater\newline
Zoomed-in video frames of the focus region:\newline
\textless video\textgreater
}
\\
\midrule
\multicolumn{3}{@{}c@{}}{\textit{Single Cue Prompts}} \\
\midrule
CQ (region) &
\parbox[t]{\linewidth}{\ttfamily
In the current view, is the \{cue\} clearly visible and large enough to follow its motion over time? Answer Yes or No.
}
&
\parbox[t]{\linewidth}{\ttfamily
Based only on visible evidence, is the person's \{cue\} visible in this view? Answer Yes or No.
}
\\

Existence (region) &
\parbox[t]{\linewidth}{\ttfamily
Is there a \{cue\} in the current view? Answer Yes or No.
}
&
\parbox[t]{\linewidth}{\ttfamily
Based only on visible evidence, can you see the person's \{cue\} in this view? Answer Yes or No.
}
\\

Latent (region) &
\parbox[t]{\linewidth}{\ttfamily
In the current view, is the \{cue\} at least partially visible but too small or blurred, so that zooming in would likely make it clearer? Answer Yes or No.
}
&
\parbox[t]{\linewidth}{\ttfamily
Is the \{cue\} visible in the current view but not yet clear enough for precise inspection, so zooming in may help? Answer Yes or No.
}
\\
\midrule
\multicolumn{3}{@{}c@{}}{\textit{Interaction Cues Prompts}} \\
\midrule
CQ (interaction) &
\parbox[t]{\linewidth}{\ttfamily
In the current view, are the \{a\} and the \{b\} both visible in the same local area, and is that local area large enough to follow possible contact or relative movement over time? Answer Yes or No.
}
&
\parbox[t]{\linewidth}{\ttfamily
Based only on visible evidence, are the person's \{a\} and \{b\} visible in the same local area of this view? Answer Yes or No.
}
\\

Existence (interaction) &
\parbox[t]{\linewidth}{\ttfamily
Are there the \{a\} and the \{b\} in the current view? Answer Yes or No.
}
&
\parbox[t]{\linewidth}{\ttfamily
Based only on visible evidence, can you see the person's \{a\} and \{b\} in the same local area of this view? Answer Yes or No.
}
\\

Latent (interaction) &
\parbox[t]{\linewidth}{\ttfamily
In the current view, are the \{a\} and the \{b\} at least partially visible in the same local area but too small or blurred, so that zooming in would likely make their contact or relative movement clearer? Answer Yes or No.
}
&
\parbox[t]{\linewidth}{\ttfamily
Are the \{a\} and the \{b\} visible in the same local area but not yet clear enough for precise inspection, so zooming in may help? Answer Yes or No.
}
\\
\bottomrule
\end{tabular}
\end{table*}

\onecolumn
\begin{center}
\small
\begin{longtable}{p{0.11\textwidth}p{0.27\textwidth}p{0.56\textwidth}}
\caption{Descriptions used for verbalizer marginalization in iMiGUE and MA-52. \label{table:appendix_descriptions}}\\
\toprule
\textbf{Dataset} & \textbf{Label} & \textbf{Description} \\
\midrule
\endfirsthead
\toprule
\textbf{Dataset} & \textbf{Label} & \textbf{Description} \\
\midrule
\endhead
\bottomrule
\endfoot

iMiGUE & Turtle neck & The movement of drawing the head and neck inward toward the shoulders, making the neck appear shortened. \\
iMiGUE & Bulging face, deep breath & The movement of taking a noticeable deep breath with visible expansion around the face and upper chest. \\
iMiGUE & Touching hat & The movement of bringing the hand close to and touching or adjusting the hat or its brim. \\
iMiGUE & Touching or scratching head & The movement of bringing the hand close to and touching or scratching the scalp or top of the head. \\
iMiGUE & Touching or scratching forehead & The movement of bringing the hand close to and touching or scratching the forehead. \\
iMiGUE & Covering face & The movement of covering a large portion of the face with one or both hands. \\
iMiGUE & Rubbing eyes & The movement of bringing the hand close to and rubbing the area around one or both eyes. \\
iMiGUE & Touching or scratching facial parts & The hand touches or scratches a local facial area such as the cheek or side of the face, not the forehead, ear, or jawline. \\
iMiGUE & Touching ears & The hand contacts or rubs the ear or earlobe at the side of the head. \\
iMiGUE & Biting nails & The movement of bringing the fingers close to the mouth and biting the nails. \\
iMiGUE & Touching jaw & The hand contacts the jawline or chin below the mouth. \\
iMiGUE & Touching or scratching neck & The hand contacts the front or side of the neck below the jaw. \\
iMiGUE & Playing or adjusting hair & The hand manipulates or rearranges visible hair strands instead of only touching the scalp or ear. \\
iMiGUE & Buckle button, pulling shirt collar, adjusting tie & The hand pinches or adjusts clothing near the collar, buttons, tie, or upper chest. \\
iMiGUE & Buckle button, pulling shirt collar, adjusting tie & The hand pinches or adjusts clothing near the collar, buttons, tie, or upper chest. \\
iMiGUE & Touching or covering suprasternal notch & The hand touches the small hollow at the base of the neck between the collarbones. \\
iMiGUE & Scratching back & The movement of reaching to and scratching the back or rear upper-body area. \\
iMiGUE & Folding arms & The movement of bringing the arms across the front of the torso in a folded or crossed posture. \\
iMiGUE & Dust offing clothes & The hand brushes or pats the clothing surface to remove dust or lint rather than adjusting the collar or tie. \\
iMiGUE & Dusting off clothes & The hand brushes or pats the clothing surface to remove dust or lint rather than adjusting the collar or tie. \\
iMiGUE & Putting arms behind body & The movement of moving one or both arms behind the torso and keeping them there. \\
iMiGUE & Moving torso & The movement of the upper body shifting, swaying, leaning, or rotating in place. \\
iMiGUE & Sitting straightly & The dynamic process of straightening the back and upper body into an upright seated posture. \\
iMiGUE & Sitting straightly & The dynamic process of straightening the back and upper body into an upright seated posture. \\
iMiGUE & Scratching or touching arms & One hand reaches across to touch or scratch the other arm or forearm. \\
iMiGUE & Rubbing or holding hands & Both hands stay in direct contact by rubbing, clasping, or holding each other, without clear finger crossing, steepling, or visible object use. \\
iMiGUE & Crossing fingers & The fingers of the two hands are visibly crossed or interlaced while the hands stay together. \\
iMiGUE & Minaret gesture & The fingertips of both hands meet in a steeple-like shape while the palms remain apart. \\
iMiGUE & Playing or manipulating objects & One or both hands visibly handle or adjust an external object held in the hands. \\
iMiGUE & Hold back arms & The movement of holding the arms back in a restrained posture close to or behind the torso. \\
iMiGUE & Head up & The movement of raising the head upward, typically as a single motion. \\
iMiGUE & Pressing lips & The movement of pressing the lips tightly together, usually without hand contact. \\
iMiGUE & Arms akimbo & The movement of placing one or both hands on the hips with the elbows angled outward. \\
iMiGUE & Shaking shoulders & The movement of the shoulders shaking or repeatedly moving up and down. \\
iMiGUE & Illustrative BLs & Communicative hand or arm gestures that accompany speech rather than self-touching or object-focused movements. \\

MA-52 & Shaking body & The movement of the upper body from side to side. \\
MA-52 & Sitting straightly & The dynamic process of straightening the back. \\
MA-52 & Shrugging & Shrugging or moving the shoulders back and forth, indicating movements involving only the shoulders. \\
MA-52 & Turning around & The movement of rotating the torso or whole body to change the facing direction. \\
MA-52 & Rising up & The preparatory movements before standing up from a seated position. \\
MA-52 & Bowing head & Lower the head slowly or abruptly, typically as a single motion. \\
MA-52 & Head up & Raise the head slowly or abruptly, typically as a single motion. \\
MA-52 & Tilting head & Tilt the head toward the shoulder. \\
MA-52 & Turning head & Turn the head backward or sideways, typically as a single motion. \\
MA-52 & Nodding & The movement of continuous up-and-down head. \\
MA-52 & Shaking head & The movement of continuous turn of the head left and right. \\
MA-52 & Scratching arms & The movement of scratching the other arm with one hand. \\
MA-52 & Playing objects & The movement of playing with or adjusting objects in hand. \\
MA-52 & Putting hands together & The movement of bringing both hands closer together until palms together. \\
MA-52 & Rubbing hands & The movement of touching and rubbing the hands together. \\
MA-52 & Pointing oneself & The movement of the hand that indicates direction and self-reference. \\
MA-52 & Clenching fist & The movement of bending the fingers into a fist, including partial clenching of the fingers. \\
MA-52 & Stretching arms & The movement of raising or extending the arm forward or sideways. \\
MA-52 & Retracting arms & The movement of extending and retracting the arm. \\
MA-52 & Waving & The movement of waving one's hand or arm. \\
MA-52 & Spreading hands & The movement of holding one's hand open with the palm facing up. \\
MA-52 & Hands touching fingers & The movement of changing from having the fingers of both hands uncrossed to interlaced. \\
MA-52 & Other finger movements & Excluding finger movements as defined within the upper limb. \\
MA-52 & Illustrative gestures & The illustrative gestures made with either the left arm or the right arm. \\
MA-52 & Shaking legs & The movement of shaking the leg up and down, either continuously or briefly. \\
MA-52 & Curling legs & The movement of sitting with one leg crossed over the other. \\
MA-52 & Spread legs & The movement of increasing the distance between the knees. \\
MA-52 & Closing legs & The movement of moving the legs from apart to together. \\
MA-52 & Crossing legs & The movement of legs gradually turning from uncrossing to crossing. \\
MA-52 & Stretching feet & The movement of extending the foot forward or sideways. \\
MA-52 & Retracting feet & Retract the foot towards the body. \\
MA-52 & Tiptoe & Place the foot sideways or rest the heel on a stool with the toe touching the ground. \\
MA-52 & Scratching or touching neck & The movement of bringing the hand close to and touching the neck area. \\
MA-52 & Scratching or touching chest & The movement of bringing the hand close to and touching the front of the upper body. \\
MA-52 & Scratching or touching back & The movement of bringing the hand close to and touching the back of the upper body. \\
MA-52 & Scratching or touching shoulder & The movement of bringing the hand close to or touching the shoulder area. \\
MA-52 & Arms akimbo & The movement of placing one or both hands on the hips. \\
MA-52 & Crossing arms & The movement of crossing one's arms over one's chest. \\
MA-52 & Playing or tidying hair & Hand movements that adjust hair on the sides or top of the head, including women arranging their hanging hair. \\
MA-52 & Scratching or touching hindbrain & Touch the back of the head or the hair on the back of the head. \\
MA-52 & Scratching or touching forehead & The movement of scratching or touching the forehead. \\
MA-52 & Scratching or touching face & The movement of touching the cheek area. \\
MA-52 & Rubbing eyes & The movement of bringing the hand close to and touching the area around the eyes. \\
MA-52 & Touching nose & The movement of bringing the hand close to and touching the nose area. \\
MA-52 & Touching ears & The movement of bringing the hand close to and touching the ear, or touching and rubbing the ear. \\
MA-52 & Covering face & The movement of covering a large part of face region, with one or both hands. \\
MA-52 & Covering mouth & The movement of covering the mouth with the hand. \\
MA-52 & Pushing glasses & The movement of bringing the hand close to and touching the glasses. \\
MA-52 & Patting legs & Move a single hand close to and touch the leg in a dynamic process. \\
MA-52 & Touching legs & The movement of touching the leg with a single hand. \\
MA-52 & Scratching legs & The movement of scratching the lower leg with the hand. \\
MA-52 & Scratching feet & The movement of scratching the foot with the hand. \\

\end{longtable}
\end{center}
\twocolumn









\end{document}